\documentclass[final]{cvpr}

\usepackage{times}
\usepackage{epsfig}
\usepackage{graphicx}
\usepackage{amsmath}
\usepackage{amssymb}

\usepackage{booktabs}

\usepackage[pagebackref=true,breaklinks=true,colorlinks,bookmarks=false]{hyperref}

\def\cvprPaperID{2469} 
\def\confYear{CVPR 2021}

\begin{document}

\title{Robust Reflection Removal with Reflection-free Flash-only Cues}


\author{Chenyang Lei \quad \quad \quad \quad Qifeng Chen\\
The Hong Kong University of Science and Technology
}

\maketitle
\pagestyle{empty}
\thispagestyle{empty}


\begin{abstract}
We propose a simple yet effective reflection-free cue for robust reflection removal from a pair of flash and ambient (no-flash) images. The reflection-free cue exploits a flash-only image obtained by subtracting the ambient image from the corresponding flash image in raw data space. The flash-only image is equivalent to an image taken in a dark environment with only a flash on. We observe that this flash-only image is visually reflection-free, and thus it can provide robust cues to infer the reflection in the ambient image. Since the flash-only image usually has artifacts, we further propose a dedicated model that not only utilizes the reflection-free cue but also avoids introducing artifacts, which helps accurately estimate reflection and transmission. Our experiments on real-world images with various types of reflection demonstrate the effectiveness of our model with reflection-free flash-only cues: our model outperforms state-of-the-art reflection removal approaches by more than 5.23dB in PSNR, 0.04 in SSIM, and 0.068 in LPIPS. Our source code and dataset are publicly available at \href{https://github.com/ChenyangLEI/flash-reflection-removal}{github.com/ChenyangLEI/flash-reflection-removal}.
\end{abstract}



\section{Introduction}
An image taken by a camera in front of a glass often contains undesirable reflection. In the process of image formation with reflection, the irradiance received by a camera can be approximately modeled as the sum of transmission and reflection. In this paper, we are interested in recovering a clear transmission image by removing reflection from the ambient image (captured image). Reflection removal is an important application in computational photography, which can highly improve image quality and pleasantness. Furthermore, computer vision algorithms can be more robust to images with reflection, as the reflection can be largely erased by a reflection removal method.




\begin{figure}
\centering
\begin{tabular}{@{}c@{\hspace{1mm}}c@{\hspace{1mm}}c@{}}
\includegraphics[width=0.323\linewidth]{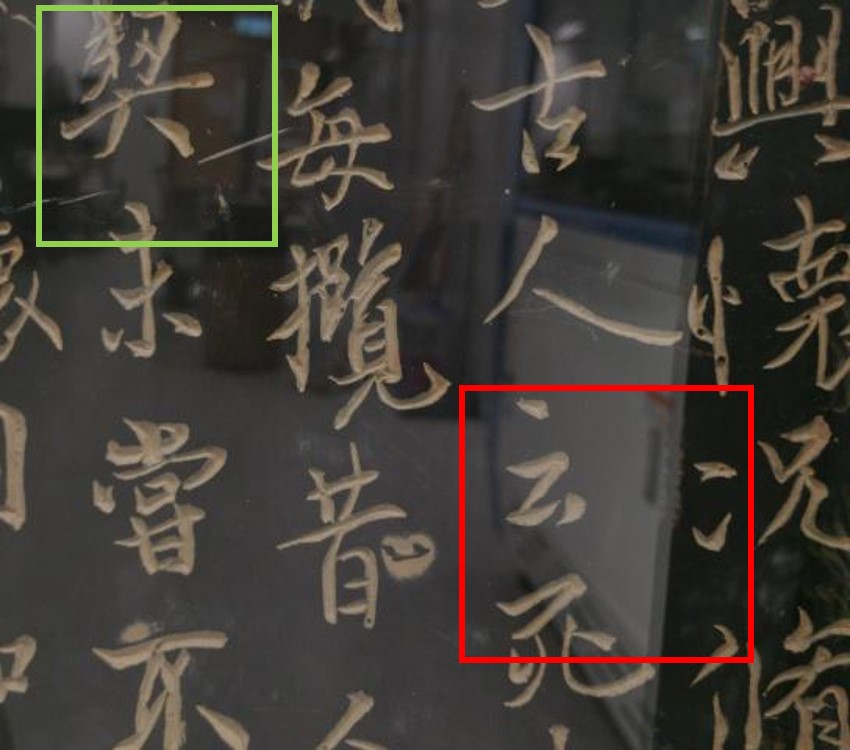}&
\includegraphics[width=0.323\linewidth]{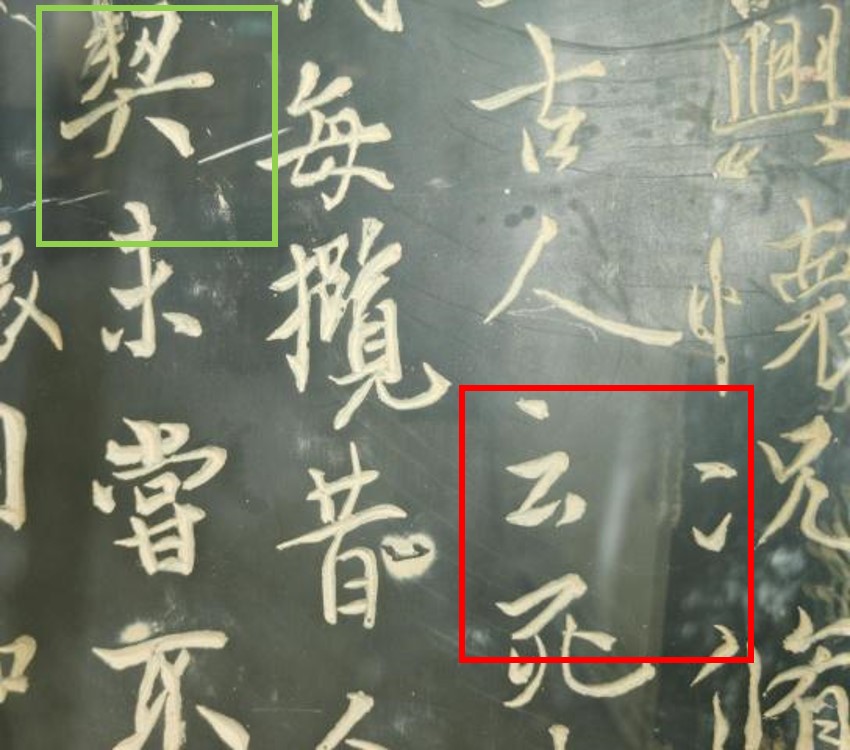}&
\includegraphics[width=0.323\linewidth]{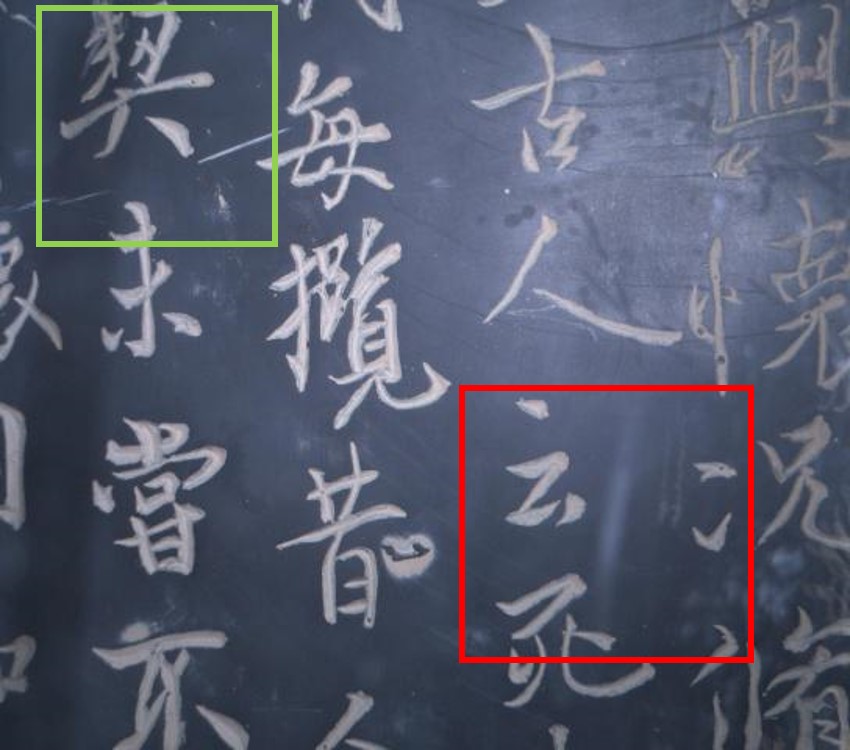}\\
Ambient image & Flash image& Flash-only image\\

\end{tabular}

\begin{tabular}{@{}c@{\hspace{1mm}}c@{}}
\includegraphics[width=0.49\linewidth]{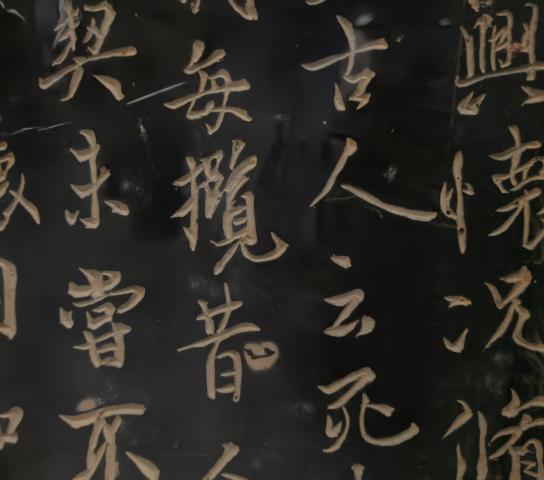}&
\includegraphics[width=0.49\linewidth]{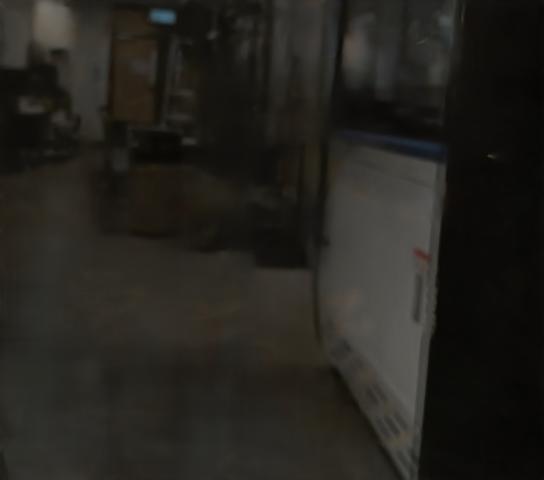}\\
 Our transmission& Our reflection \\

\end{tabular}
\vspace{1mm}
\caption{A reflection-free flash-only image is computed from a pair of ambient/flash images to help remove reflection. Our transmission image does not absorb the artifacts in the flash-only image.}
\label{fig:Teaser}
\end{figure}

Reflection removal is challenging because the reflection component is usually unknown. Since both reflection and transmission are natural images, it is hard to distinguish between reflection and transmission from an input image. Therefore, many methods adopt various assumptions on the appearance of reflection for reflection removal. For example, some single image-based methods~\cite{Arvanitopoulos_2017_CVPR, Yang_2019_CVPR} assume that the reflection is not in-focus and blurry. The ghosting cue~\cite{shih2015reflection_ghost} is another assumption that holds when the glass is thick. However, reflection in real-world images is diverse, and these assumptions do not necessarily hold~\cite{Lei_2020_CVPR,wan2017benchmarking}. As a result, existing methods cannot perfectly remove reflection with diverse appearance from real-world images~\cite{Lei_2020_CVPR}.

We propose a novel \emph{reflection-free flash-only cue} that facilitates inferring the reflection in an ambient image. This cue is robust since it is independent of the appearance and strength of reflection, unlike assumptions adopted in previous single image reflection removal methods~\cite{shih2015reflection_ghost, Yang_2019_CVPR} or flash-based methods~\cite{chang2020siamese}. The reflection-free cue is based on a physics-based phenomenon of an image obtained by subtracting an ambient image from the corresponding flash image (in raw data space). This flash-only image is equivalent to an image captured under the flash-only illumination: the environment is completely dark, and a single flash is the sole light source. A key observation is that the reflection is invisible in the flash-only image. 

While flash-only images provide reflection-free cues to distinguish reflection, they also have weaknesses. For instance, in Fig.~\ref{fig:Teaser}, we can observe artifacts (e.g., color distortion, illuminated dust) due to uneven flash illumination, occlusions, and other reasons. These artifacts prevent us from obtaining a high-quality transmission easily.

To utilize the reflection-free cue and avoid introducing the flash-only image artifacts, we design a dedicated architecture for obtaining high-quality transmission. Specifically, we first estimate a reflection image instead of a transmission image. Then, to further avoid introducing artifacts in the flash-only image, only the input ambient image and the estimated reflection are given to the second network that estimates the transmission. 

Combining our dedicated architecture with the reflection-free cue, we can robustly and accurately remove various kinds of reflection to recover the underlying transmission image. Although we need an extra flash image compared with single image methods, a flash/no-flash image pair can be captured with a single shutter-press using customized software, as shown in Fig.~\ref{fig:App}. Hence, general users can easily apply our method for robust reflection removal. In summary, our contributions are as follows:
\begin{itemize}
    \item We propose a novel cue - the reflection-free flash-only cue that makes distinguishing reflections simpler for reflection removal. This cue is robust since it is independent of the appearance and strength of reflection. 
    
    \item We propose a dedicated framework that can avoid introducing artifacts of flash-only images while utilizing reflection-free cues. We improve more than 5.23dB in PSNR, 0.04 in SSIM, and 0.068 in LPIPS on a real-world dataset compared with state-of-the-art methods. 

    \item We construct the first dataset that contains both raw data and RGB data for flash-based reflection removal.

\end{itemize}

\section{Related Work}
\subsection{Reflection Removal}

\textbf{Single image reflection removal.} 
In single image reflection removal, the defocused reflection assumption and ghosting cue are commonly used. The defocused reflection assumption means that reflections are not in focus. Hence, prior work can assume they are more blurry compared with the transmission. Following this assumption, learning-based methods~\cite{fan2017generic,zhang2018single} can synthesize abundant data for training, and non-learning based methods can suppress the reflection based on image gradient~\cite{Arvanitopoulos_2017_CVPR, Yang_2019_CVPR}. The ghosting cue means multiple reflections are visible on the glass~\cite{shih2015reflection_ghost}. However, the ghosting cue only exists when the glass is thick. Hence, algorithms that are based on ghosting cue might fail on the thin glass. 

There are many attempts to relax assumptions of reflection. Wei et al.~\cite{Wen_2019_CVPR_Linear}
and Ma et al.~\cite{Ma_2019_ICCV} use generative adversarial networks~\cite{DBLP:conf/nips/GoodfellowGAN14} to synthesize realistic reflection under the guidance of real-world reflections. Kim et al.~\cite{Kim_2020_CVPR} propose a physics-based method to render the reflection and mixed image, which improves the quality of training data a lot. Also, Zhang et al.~\cite{zhang2018single}, Wei et al.~\cite{wei2019single_ERR}, and Li et al.~\cite{Li_2020_CVPR} collected real-world data for improving the quality of training data. However, as reported by Lei et al.~\cite{Lei_2020_CVPR}, these methods~\cite{Arvanitopoulos_2017_CVPR,fan2017generic, Yang_2019_CVPR} are still far
from perfectly removing reflections for diverse real-world data.

\textbf{Multiple images reflection removal.}
Some reflection removal methods utilize the motion cue of reflection and transmission in multiple images for reflection removal~\cite{guo2014robust,han2017reflection,li2013exploiting,liu2020learning,DBLP:conf/mm/Sun16RR,DBLP:journals/tog/xue2015computational}. In these motion-based methods, SIFT-flow~\cite{li2013exploiting}, homography~\cite{guo2014robust} and optical flow~\cite{liu2020learning,DBLP:journals/tog/xue2015computational} are used to find correspondences among multiple images to distinguish reflection and transmission. However, taking images with different motion cost more effort, and some assumptions are required (e.g., all pixels in transmission must appear in at least one image~\cite{DBLP:journals/tog/xue2015computational}). Polarization is also used in reflection removal to achieve great performance~\cite{Fraid1999,kong14pami,lyu2019reflection,nayar1997separation,eccv2018/Wieschollek, Schechner1999PolarizationbasedDO}. The inputs are usually images through various polarizers, which contain polarization information of light. Since polarization of reflection and transmission is usually different, it can be used to distinguish them. However, a polarizer is usually required to be shifted to take images, which is complicated. Recently, a camera~\cite{Li_2020_CVPR,Li_eccv20_PolarRR} that can take several polarization images appears but this kind of camera is yet to be widely used.

\begin{figure*}[t]
\centering
\begin{tabular}{@{}c@{}}
\includegraphics[width=1.0\linewidth]{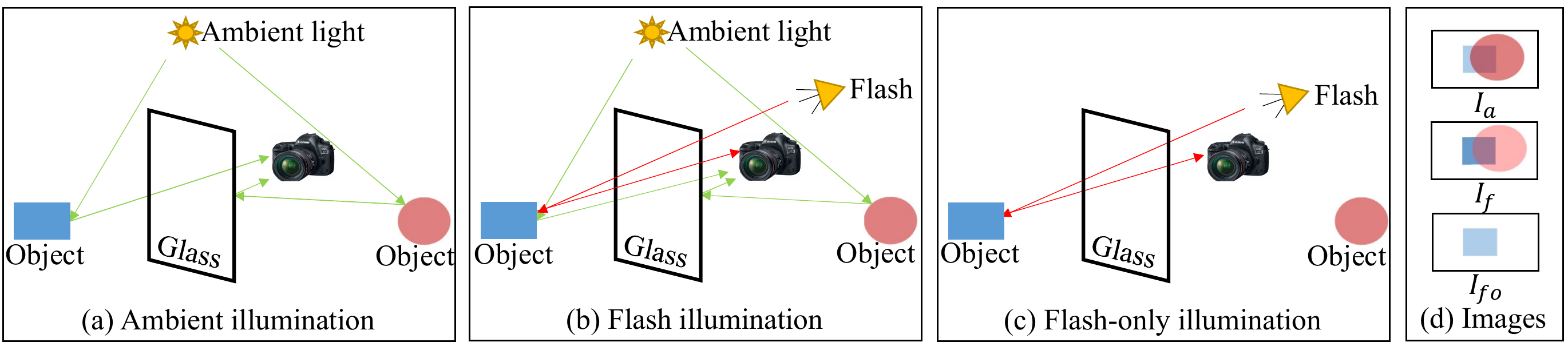}\\
\end{tabular}
\caption{An illustration model of the reflection-free cue. Since objects in reflection cannot \emph{directly} receive flash and reflected flash from glass is often weak, flash-only images are visually reflection-free. Note that the flash-only image $I_{fo}$ is obtained from $I_a$ and $I_f$.}
\label{fig:Illustration.}
\end{figure*}

\begin{figure}[t]
\centering
\begin{tabular}{@{}c@{\hspace{1mm}}c@{\hspace{1mm}}c@{\hspace{1mm}}c@{}}
\rotatebox{90}{ \hspace{1.5mm}  Sample 1}&
\includegraphics[width=0.31\linewidth]{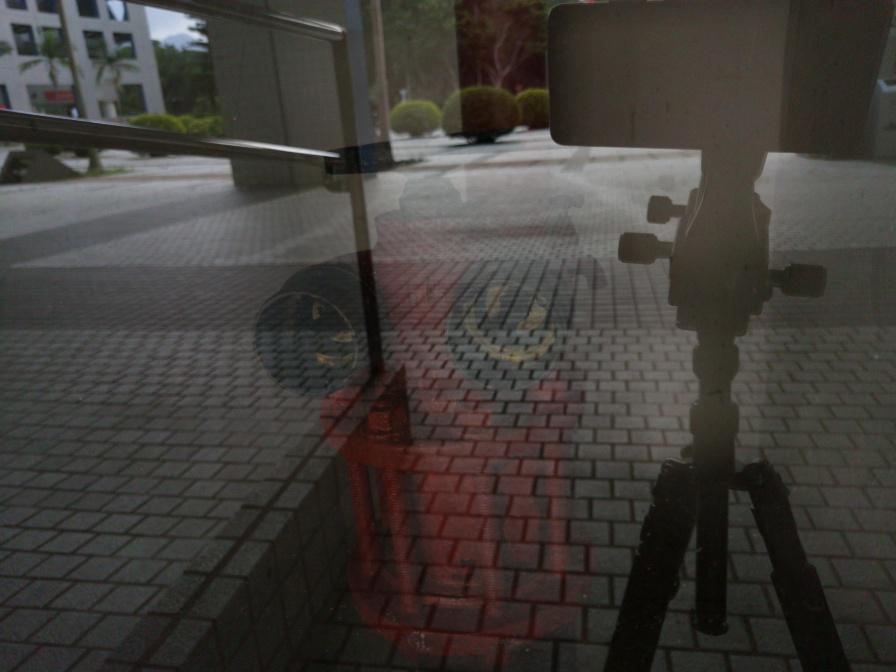}&
\includegraphics[width=0.31\linewidth]{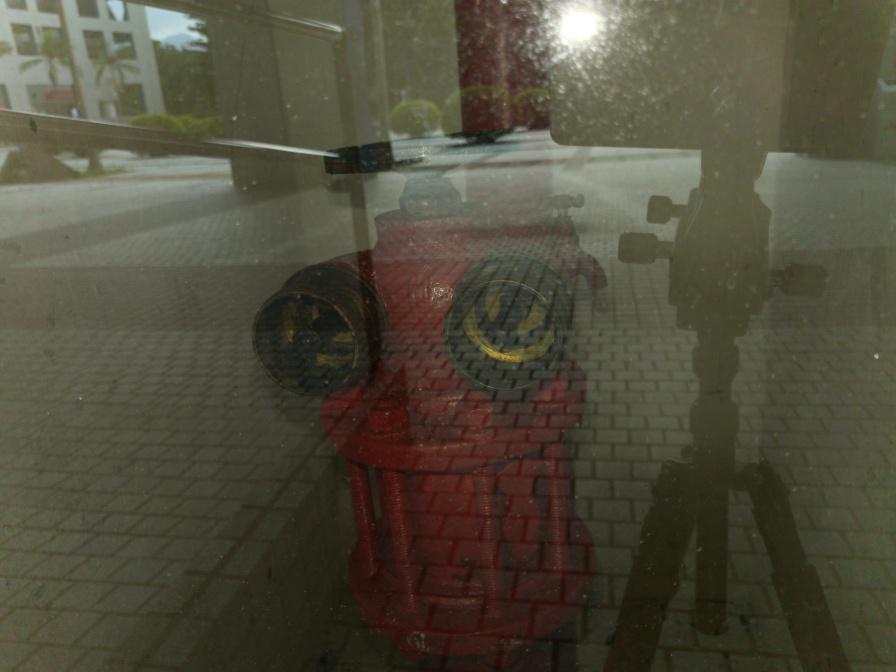}&
\includegraphics[width=0.31\linewidth]{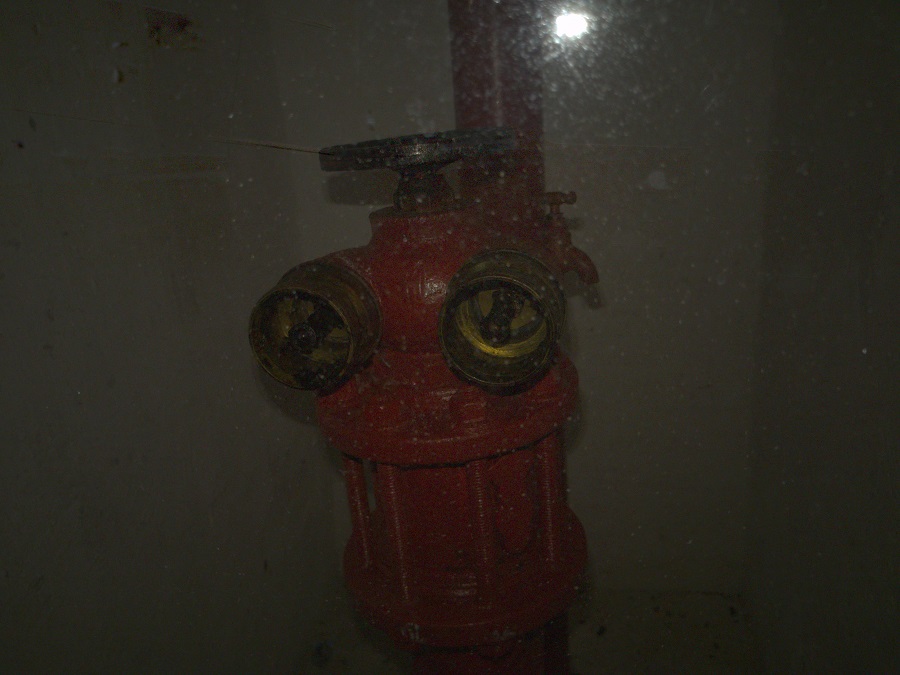}\\


\rotatebox{90}{ \hspace{1.5mm} Sample 2 }&
\includegraphics[width=0.31\linewidth]{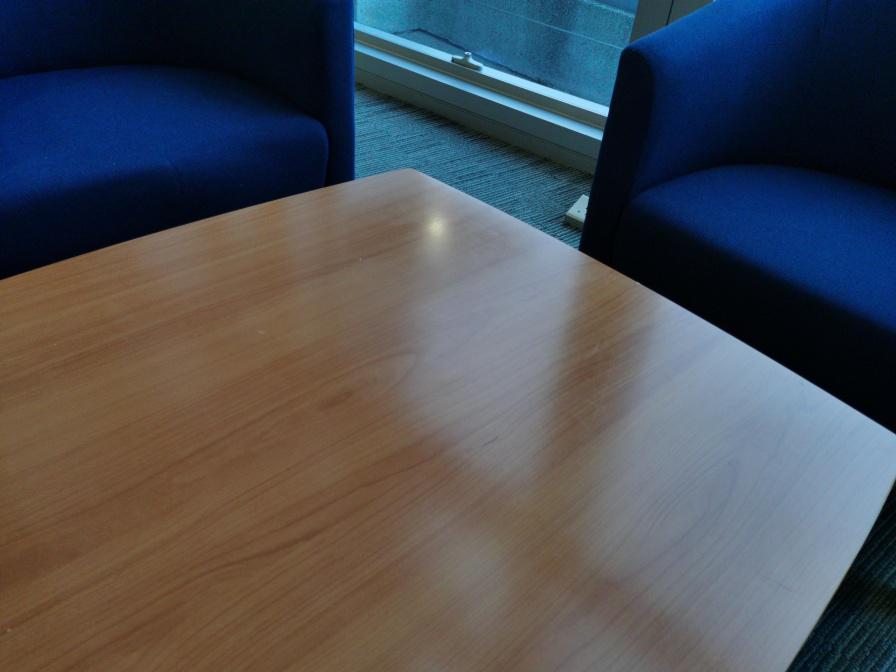}&
\includegraphics[width=0.31\linewidth]{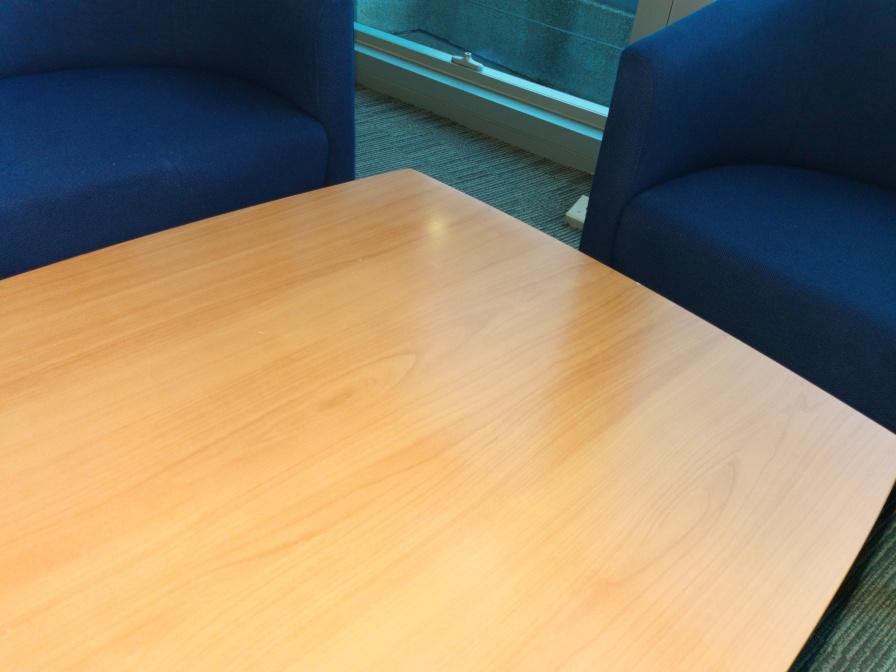}&
\includegraphics[width=0.31\linewidth]{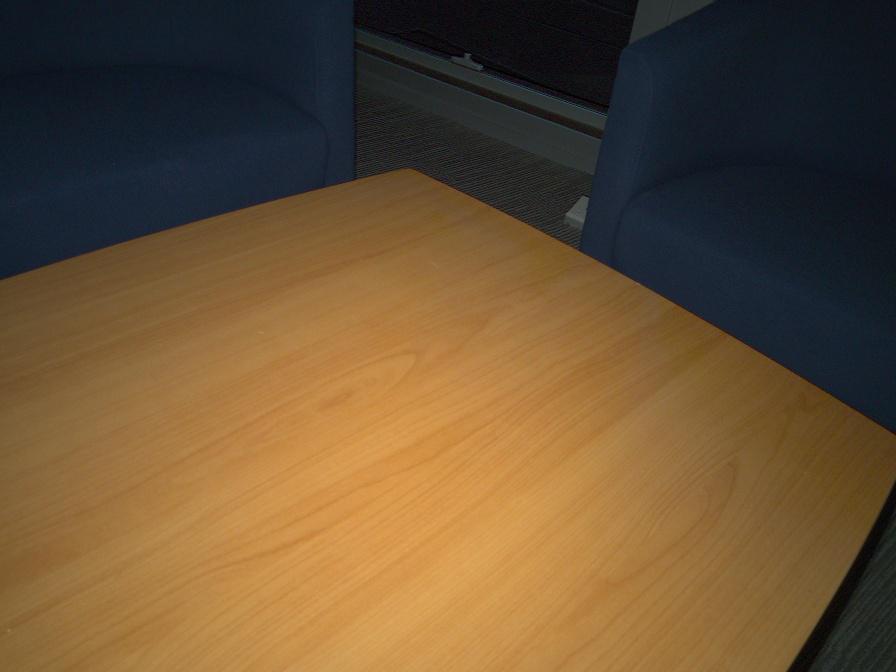}\\

 &Ambient images & Flash images  & Flash-only images \\

\end{tabular}
\caption{Examples of reflection-free flash-only images. Flash-only images are visually reflection-free but have many artifacts.}
\label{fig:pureflash}
\end{figure}

\textbf{Flash-based reflection removal.}
Various properties of a pair of flash/ambient images are adopted in previous work~\cite{agrawal2005removing_flash, chang2020siamese}. Agrawal et al.~\cite{agrawal2005removing_flash} claim that gradient orientations are consistent in the image pair, assuming that depth edges, shadows, and highlights are few. 
However, they cannot generate reasonable results for undesirable regions (e.g., shadows, specular reflection), and their results tend to be over-smooth. SDN~\cite{chang2020siamese} utilize the assumption that reflection can be obviously suppressed by flash, but the suppression effect is sensitive to the strength of reflection: when reflection is strong, the suppression is no longer effective.

\subsection{Flash Photography}
Flash images are used in various tasks. Petschnigg et al.~\cite{DBLP:journals/tog/PetschniggSACHT04} use the flash image for denoising, detail transfer, etc. Drew et al.~\cite{drew2006removing} use the flash-only image for shadow removal.
Sun et al.~\cite{sun2005flashmatting} observe that the change of intensity is different for near objects and background in the flash-only image and apply it to image matting. Cao et al.~\cite{cao2020stereoscopic} use the flash-only image for shape and albedo recovery under the assumption of the Lambertian model.


\section{Reflection-free Flash-only Cues}
\label{sec:flashonly}
\textbf{Flash-only images.}
Let $\{I_a, I_{f}\}$/$\{I_a^{raw}, I_{f}^{raw}\}$ be the RGB/raw images under ambient and flash illuminations. A flash-only image $I_{fo}^{raw}$ can be computed from $I_a^{raw}$ and $I_{f}^{raw}$. Since the flash image is the sum of ambient image and flash-only image for a linear response camera in linear space~\cite{DBLP:journals/tog/PetschniggSACHT04}, we can obtain the flash-only image through:
\begin{align}
\label{eq:pureflash}
I_a^{raw} &= R_{a}^{raw} + T_a^{raw}, \\    
I_{f}^{raw} &=R_{a}^{raw} + T_a^{raw} + R_{fo}^{raw} + T_{fo}^{raw}, \\
I_{fo}^{raw} &= I_{f}^{raw} - I_a^{raw}=R_{fo}^{raw} + T_{fo}^{raw},
\end{align}
where $R$ and $T$ are the reflection and transmission. For simplicity, we also use $I^{raw}$ to denote the image after linearization. Reflection-free cues exist in flash-only images. The flash-only image is equivalent to an image captured in a completely dark environment, and the flash is the sole light source, as shown in Fig.~\ref{fig:Illustration.}(b). Note that $I_{fo}^{raw}$ is invariant to different ambient illuminations as long as $I_a^{raw}$ and $I_f^{raw}$ do not have saturated pixels.


\textbf{Reflection-free cues.} 
The reflection-free cue denotes a physics-based phenomenon: reflections of ambient image $R_a^{raw}$ are invisible in the flash-only image. Besides, we have $R_{fo}^{raw}\approx \mathbf{0}$ in most cases. Fig.~\ref{fig:Illustration.} shows an illustration of this phenomenon. Reflections exist in ambient images because objects in the reflection receive ambient light and then reflect it to the camera through the glass. In Fig.~\ref{fig:Illustration.}(b), objects in reflection do not directly receive light from the flash. Besides, since reflectance of glass is mostly much weaker than transmittance, the reflected flash is almost negligible (please check the supplement for detailed analysis). Hence, objects in reflection are barely illuminated and reflections do not appear in flash-only images. 


To verify reflection-free cues, we capture pairs of ambient and flash images under different illumination and scenes, and we compute $I_{fo}^{raw}$ following Eq.~\ref{eq:pureflash}. As shown in Fig.~\ref{fig:pureflash}, reflections disappear in flash-only images, even when reflections are strong. Also, the 2nd example shows this cue is valid not only for semi-reflecting surfaces.

\begin{figure*}
\centering
\begin{tabular}{@{}c@{}}
\includegraphics[width=1.0\linewidth]{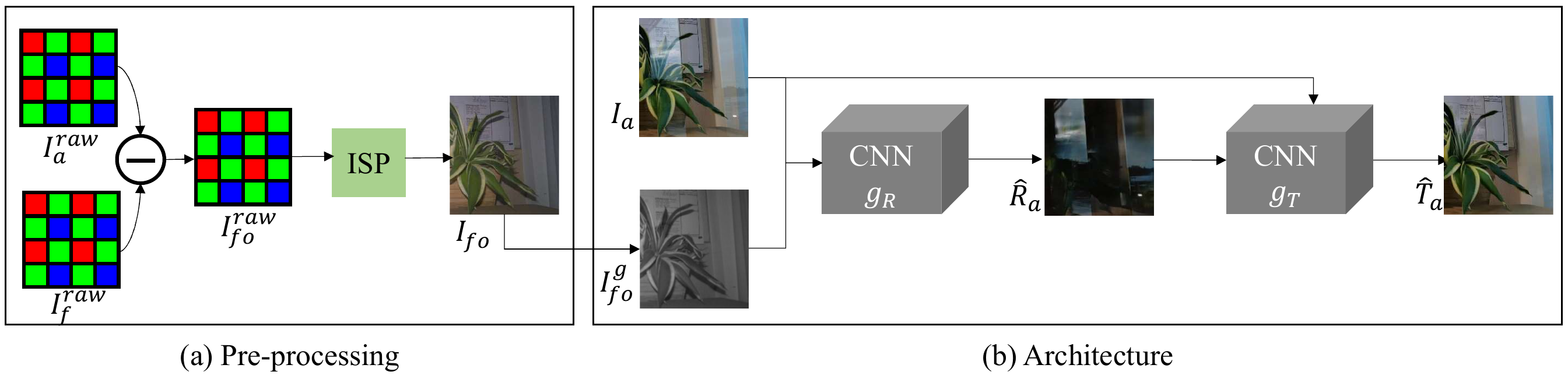}
\end{tabular}
\caption{The overall architecture of our approach. We compute the $I_{fo}$ from $I_f^{raw}$ and $I_a^{raw}$. Then, our dedicated architecture estimates the reflection first to avoid absorbing artifacts of flash-only images. Finally, the transmission is estimated with the guidance of reflection. }
\label{fig:Architecture}
\end{figure*}


\textbf{Undesirable artifacts.} 
Although flash-only images are visually reflection-free, they usually have undesirable artifacts, as shown in Fig.~\ref{fig:pureflash}. We can analyze reasons of degradation formally from {flash-only radiance} by Eq.~\ref{eq:radiance}~\cite{kajiya1986rendering}:
\begin{align}
\label{eq:radiance}
    L_o^{fo} &= \int_{\Omega}f_r({\omega}_i,\omega_o)L_i(\omega_i)(\omega_i \cdot \mathbf{n})d \omega_i,\\
    \label{eq:Falloff}
    L_{i,d}^{fo}(w_i)& = \frac{L^{fo}(w_i)}{d^2}, 
\end{align}
where $L_o$ and $L_i$ are the radiance of outgoing and incident light, $\omega_i$ and $\omega_o$ are the light direction of outgoing and incident light, $f_r$ is the bidirectional reflectance distribution function (BRDF), $\mathbf{n}$ is the surface normal and $\Omega$ is the hemisphere. Flash-only images can contain the following artifacts that require to be resolved:

(1) Color distortion usually appears since the flash light $L^{fo}$ is different from ambient illumination. Similarly, the shading also changes since light direction $w_i$ is different.

(2) Uneven illumination is a common problem due to irradiance falloff in Eq.~\ref{eq:Falloff}, irradiance (and thus radiance $L_{i,d}^{fo}$) is different due to different distance $d$ to the flash.

(3) New shadows are brought by occlusion. 

(4) Highlights caused by flash might appear on the glass.

(5) If the glass is dirty, dust can be illuminated on the glass, as shown in 1st example in Fig.~\ref{fig:pureflash}.

\section{Method}



Given an ambient image $I_a^{raw}$ and a flash image $I_f^{raw}$ in raw space, our approach aims to estimate the transmission $T_a$ under ambient illumination. In Fig.~\ref{fig:Architecture}(a), we first take raw images $I_a^{raw}, I_f^{raw}$ and implement pre-processing to obtain the RGB flash-only image $I_{fo}$, as introduced Sec.~\ref{sec:data_prepare}. Then, our dedicated architecture in Sec.~\ref{sec:Architecture} takes RGB images $I_a, I_{fo}$ as input to remove the reflection. 

\subsection{Pre-processing}
\label{sec:data_prepare}

We first capture two raw images $I_a^{raw}$ and $I_f^{raw}$ through pipeline in Sec.~\ref{subsec:Dataset}. Given $I_a^{raw}$ and $I_f^{raw}$, we implement the following pipeline to obtain RGB images:

1) Subtraction. We first implement linearization to convert images to linear space using the black-level and white-level information from the metadata. After this step, the range of each pixel is transferred to $[0, 1]$. Then the flash-only image is obtained through Eq.~\ref{eq:pureflash} since the linearity between pixel values and physical light is preserved well. At last, the flash-only image is converted back to raw space using the black-level and white-level information.

2) Image signal processing (ISP). We implement a regular ISP~\cite{2016ISP} that includes linearization, demosaiced, white balance, color correction, and gamma correction to convert raw images to RGB images using the original metadata of images. We adopt the metadata of $I_a^{raw}$ to process $I_{fo}^{raw}$ since it is obtained by $I_a^{raw}, I_f^{raw}$, and no metadata is available. Note that the white balance of $I_{fo}$ is usually not as good as $I_a$ since it does not have its own metadata. For $I_a$ at test time, we can use our ISP to obtain the sRGB image or use the original sRGB image processed by the camera's ISP. A learning-based ISP~\cite{Chen2018SID,xing2021cvpr,Zamir2020CycleISP} can also be used here.

\subsection{Architecture}
\label{sec:Architecture}
\label{subsec:basemodel}
\begin{figure}[t!]
\centering
\begin{tabular}{@{}c@{\hspace{1mm}}c@{\hspace{1mm}}c@{\hspace{1mm}}c@{}}
\rotatebox{90}{\hspace{4mm} \small Input}&
\includegraphics[width=0.305\linewidth]{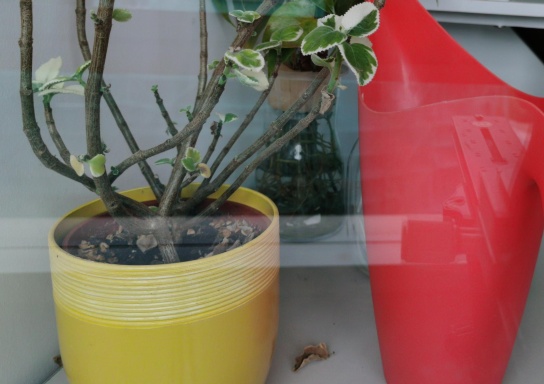}&
\includegraphics[width=0.305\linewidth]{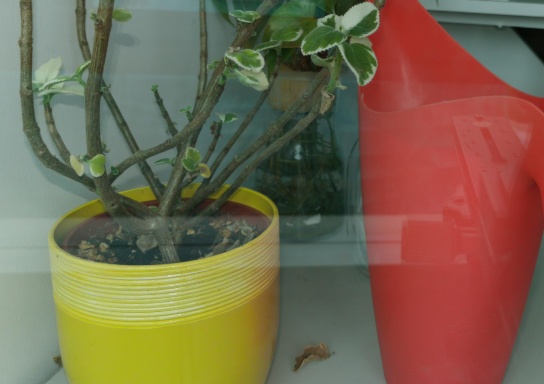}&
\includegraphics[width=0.305\linewidth]{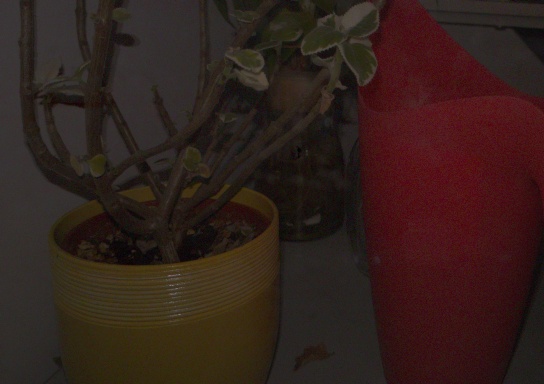}\\
&(a) Input $I_a$ &(b) Input $I_f$ & (c) Processed $I_{fo}$\\
\rotatebox{90}{\hspace{4mm} \small Our $\hat T$}&
\includegraphics[width=0.305\linewidth]{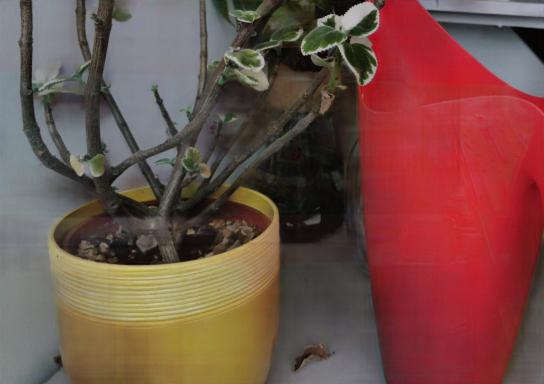}&
\includegraphics[width=0.305\linewidth]{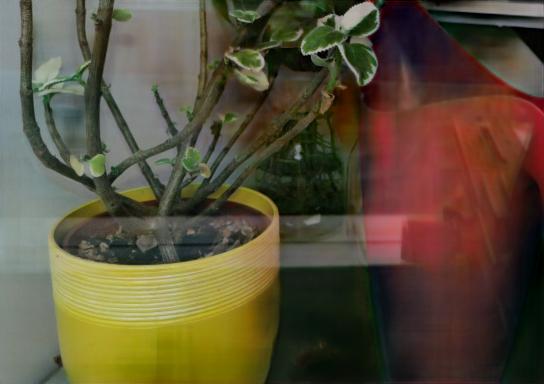}&
\includegraphics[width=0.305\linewidth]{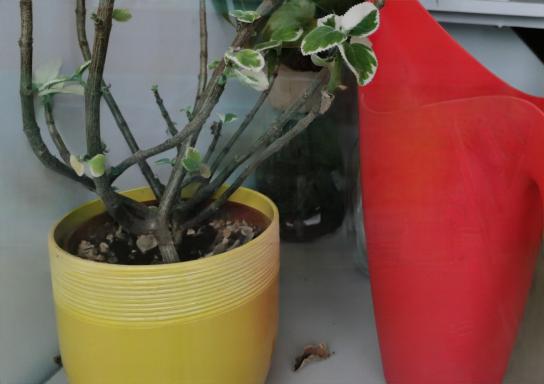}\\
&{(d) $g_B$, $I_{fo}$}   & (e) {$g_R+g_T$, $I_f$ }&  
{(f) $g_R+g_T$, $I_{fo}$}\\
\end{tabular}
\vspace{1mm}
\caption{Qualitative comparison among multiple implementations. Combining the reflection-free $I_{fo}$ with our dedicated architecture achieves the best performance.}

\label{fig:AblationStudy}
\end{figure}

As shown in Fig.~\ref{fig:pureflash}, the reflection in $I_a$ does not exist in $I_{fo}$. Except for the artifacts and color distortion, the $I_{fo}$ is quite similar to our target transmission $T_a$. Hence, we first try to use a network to directly estimate transmission from $I_a$ and $I_{fo}$, which we denote as base model $g_B$ (note that this model $g_B$ is \emph{not our final model}):
\begin{align}
    \hat T_B = g_B(I_a,I_{fo};
    \theta_B),
\end{align}
where $\theta_B$ is the parameters of network $g_B$. However, we observe that although this model can correctly remove various types of reflections$R_a$, the estimated $\hat T_B$ has undesirable artifacts, especially for the area that contains shadows, highlight in flash-only images. Also, the color might be closer to $I_{fo}$ in some area (i.e., color distortion), as shown in Fig.~\ref{fig:AblationStudy}(d). 


We argue that: since the transmission component is the intersection of $I_{fo}$ and $I_a$, the network tends to fuse $I_{fo}$ and $I_{a}$ to obtain the estimated transmission, and artifacts of $I_{fo}$ are inevitably fused too. 

\subsubsection{Reflection-pass network}


To solve the drawback of $g_B$, we only estimate the reflection first instead of directly estimating $T_a$. Stated in another way, only the reflection passes the first network. As reflections $R_a$ only exist in $I_a$, it must be extracted from $I_a$ and avoid introducing artifacts of $I_{fo}$. On the other hand, since there is no reflection in the flash-only image $I_{fo}$, it can provide strong guidance for reflection estimation. Specifically, we first convert the flash-only image to grayscale image $I_{fo}^g$ to avoid the influence of color distortion. In practice, we find that the grayscale flash-only image can provide enough structure information for estimating the reflection. Then, $I_a, I_{fo}^g$ are concatenated as input to the network $g_R$:
\begin{align}
    \hat R_a &= g_R(I_a,I_{fo}^g; \theta_R), \\
    L_R(R_a, \hat R_a) &= ||R_a - \hat R_a||_2^2,    
\end{align}
where $\theta_R$ is the parameters of network $g_R$. We adopt the L2 loss for training $g_R$.

\subsubsection{Reflection-guided transmission estimation}
The effectiveness of adopting reflection as guidance has been proven in previous work~\cite{Lei_2020_CVPR, Li_2020_CVPR,eccv18refrmv_BDN}. Hence, we directly use the estimated reflection and the ambient image to estimate the transmission. Note that the flash-only image $I_{fo}$ is \emph{not input} to $g_T$ to avoid introducing artifacts. The transmission $\hat{T_a}$ is then estimated: 
\begin{align}
    \hat T_a &= g_T(I_a, \hat R_a; \theta_T), 
\end{align}
where $\theta_T$ is the parameters of $g_T$. We also adopt a L2 loss for training $g_T$. As shown in Fig.~\ref{fig:AblationStudy}(f), the result of $g_R + g_T$ does not contain obvious artifact (e.g., color distortion), which is much better than the result of $g_B$ in Fig.~\ref{fig:AblationStudy}(d).

\emph{Discussion} One might argue that reflection-free flash-only images (Eq.~\ref{eq:pureflash}) can be learned implicitly using a large amount of data. However, note that the linearity does not exist in RGB images after non-linear ISP operation. Using the same training setting, replacing $I_{fo}$ with $I_{f}$ can lead to artifacts on strong reflection, as shown in Fig.~\ref{fig:AblationStudy}(e).


\subsubsection{Implementation details} 
We train for 150 epochs with batch size 1 on an Nvidia RTX 2080 Ti GPU. We use the Adam optimizer~\cite{DBLP:journals/corr/KingmaB14} to update the weights with an initial learning rate of $10^{-4}$. The plain U-Net~\cite{Ronneberger2015} is used for the two networks (with trivial modification~\cite{lei2020dvp}). The two networks $g_R$ and $g_T$ are trained simultaneously. We implement random cropping for images with more than 640,000 pixels. For the other images, the network is trained on complete images rather than patches.

\begin{figure}[t!]
\centering
\begin{tabular}{@{}c@{\hspace{2mm}}c@{}}
\includegraphics[width=0.36\linewidth]{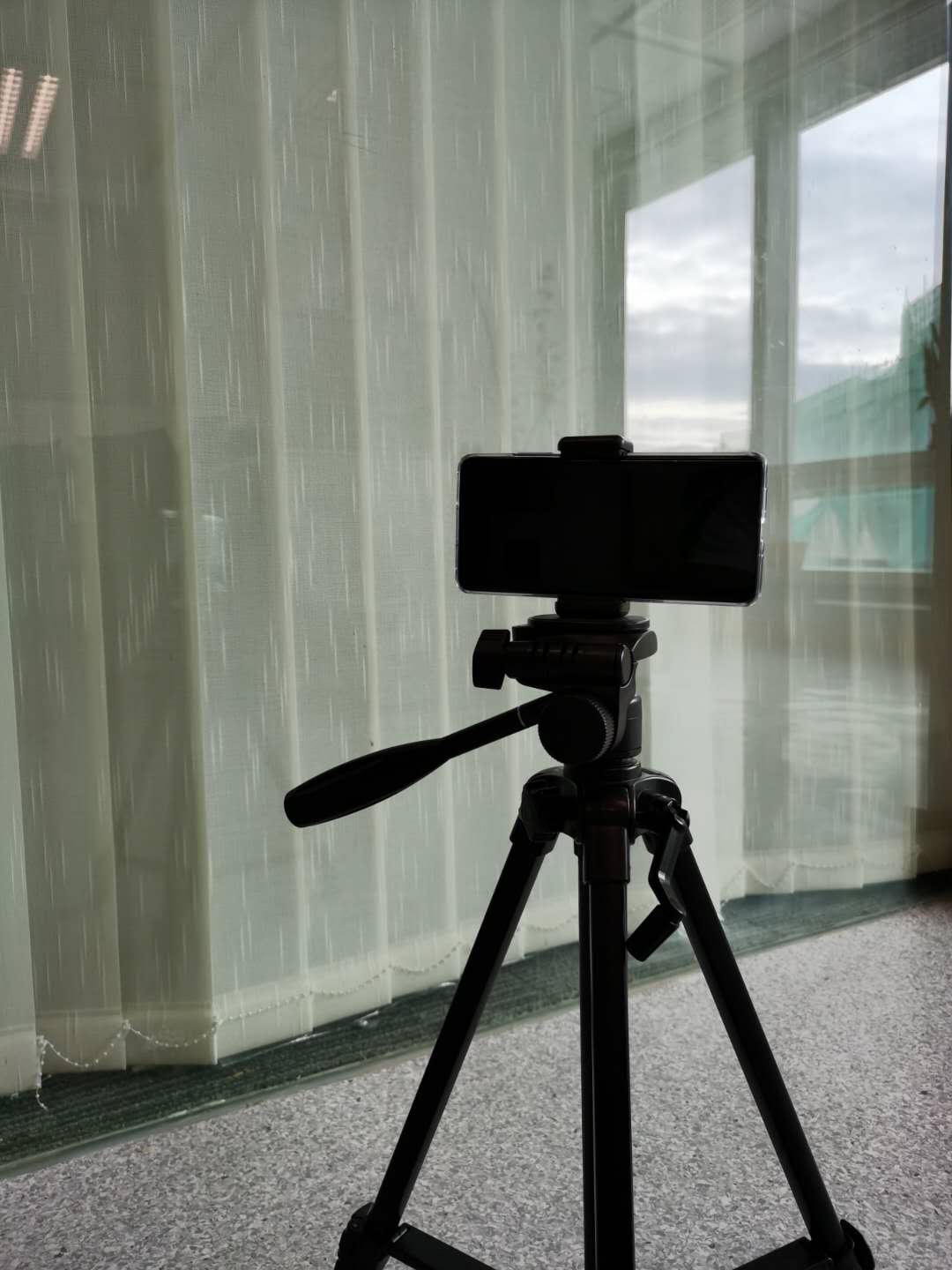}&

\includegraphics[width=0.590\linewidth]{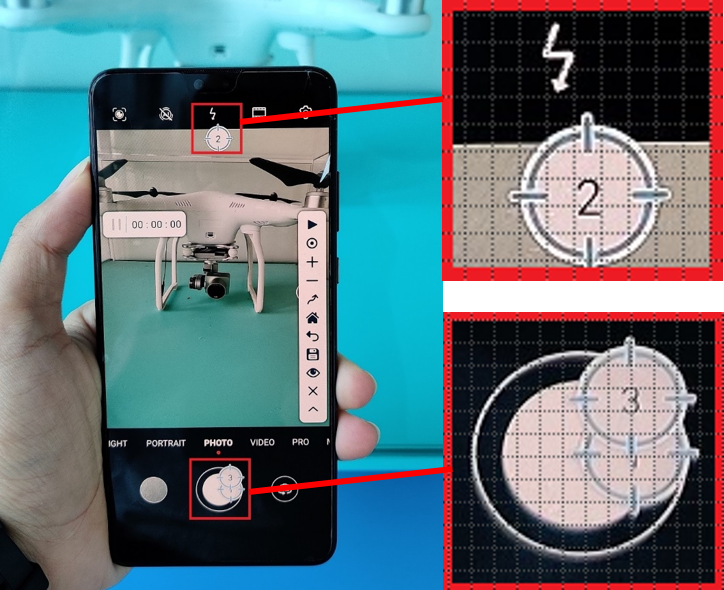}\\
(a) & (b) \\

\end{tabular}
\vspace{1mm}
\caption{A picture of (a) our data acquisition setup for constructing the dataset and (b) a real application. In (b), a user only needs to click a button to capture a pair of flash/ambient images.}
\label{fig:App}
\end{figure}

\begin{table*}[t]
\small
\centering
\renewcommand{\arraystretch}{1.2}

\begin{tabular}{cc@{\hspace{5.1mm}}c@{\hspace{5.1mm}}c@{\hspace{5.1mm}}c@{\hspace{5.1mm}}c@{\hspace{5.1mm}}c@{\hspace{5.1mm}}c@{\hspace{5.1mm}}c@{\hspace{5.1mm}}c@{\hspace{5.1mm}}}

\toprule[1pt]

     & Input $I_a$& {Zhang}  & BDN  & Wei et al. & Kim et al. & Li et al. & {Agrawal} & {SDN} & Ours\\ 


              &  & et al.~\cite{zhang2018single} & ~\cite{eccv18refrmv_BDN} & ~\cite{wei2019single_ERR} & ~\cite{Kim_2020_CVPR} &~\cite{Li_2020_CVPR} &   et al.~\cite{agrawal2005removing_flash} & ~\cite{chang2020siamese} & \\ 


\midrule

 \#Input images& 1 & 1 & 1  & 1   &  1   & 1 & 2 & 2 & 2 \\ 
\midrule
 PSNR$\uparrow$ & 22.72 & 23.76 & 21.41 & 23.89 & 21.67 & \underline{24.53} & 23.13 &                       22.63 & \textbf{29.76} \\ 
    SSIM$\uparrow$ & 0.874 & 0.873 & 0.802 & 0.864 & 0.821  &\underline{0.890} & 0.853 &             0.827 & \textbf{0.930} \\ 
 LPIPS$\downarrow$ & \underline{0.205} & 0.242 & 0.410 & 0.238 & 0.298  &{0.224} & 0.251 &             0.269 & \textbf{0.156} \\ 

\bottomrule[1pt]
\end{tabular}
\vspace{1mm}
\caption{Quantitative comparison results among our method and previous methods on a real-world dataset. }
\label{table:Metrics}
\end{table*}
\begin{figure*}[t!]
\centering
\begin{tabular}{@{}c@{\hspace{1mm}}c@{\hspace{1mm}}c@{\hspace{1mm}}c@{\hspace{1mm}}c@{\hspace{1mm}}c@{}}

\includegraphics[width=0.1601\linewidth]{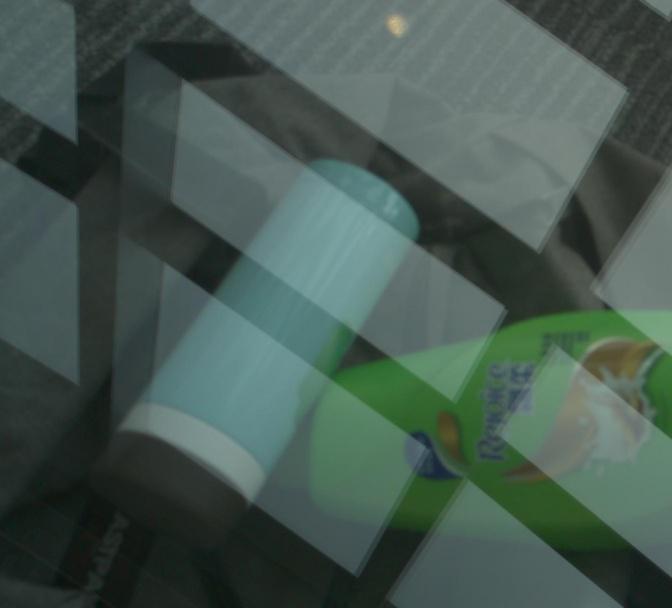}&
\includegraphics[width=0.1601\linewidth]{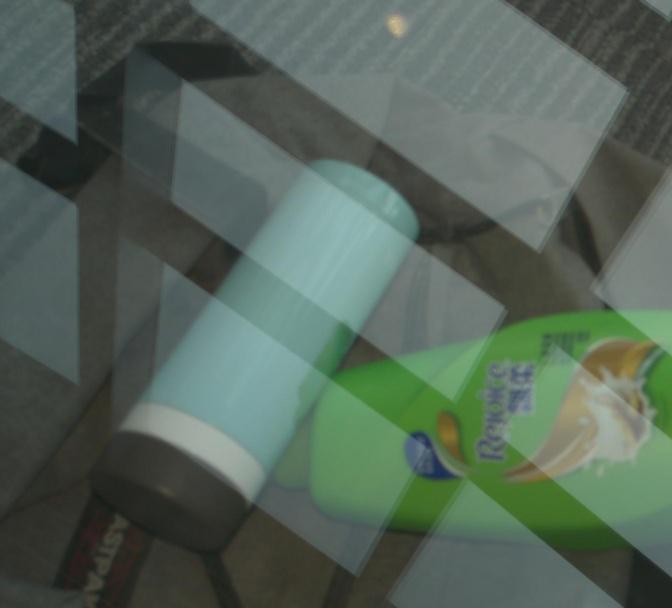}&
\includegraphics[width=0.1601\linewidth]{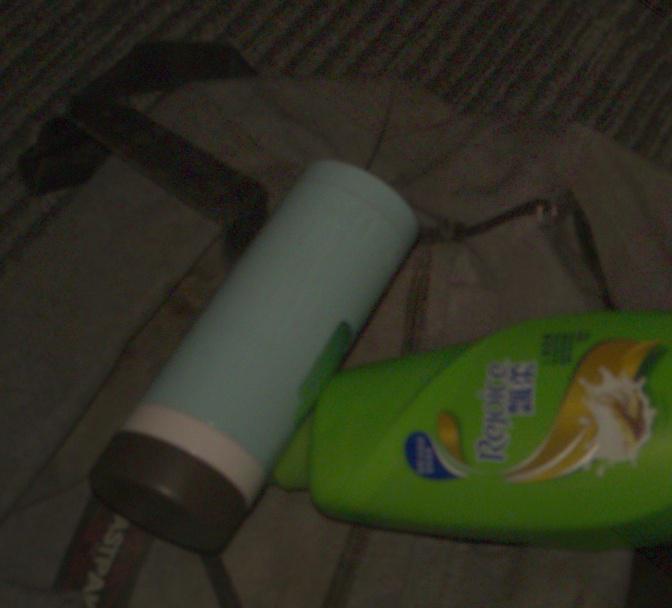}&

\includegraphics[width=0.1601\linewidth]{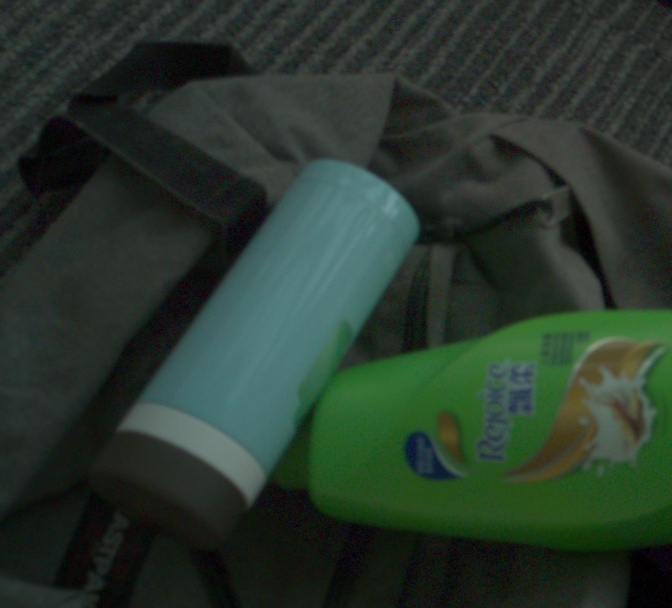}&
\includegraphics[width=0.1601\linewidth]{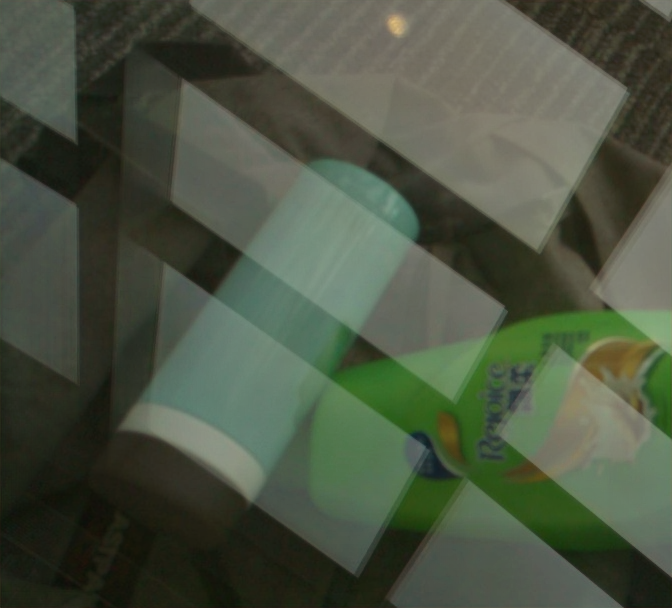}&
\includegraphics[width=0.1601\linewidth]{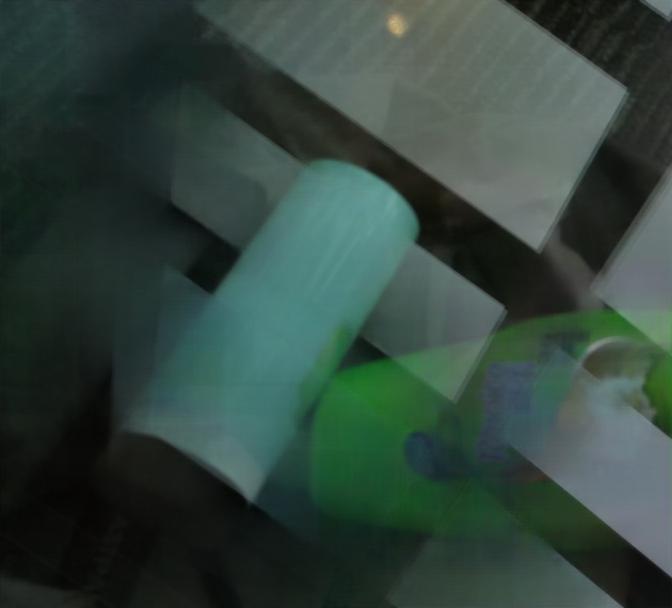}\\
Input $I_a$  &  Input $I_f$  & Processed $I_{fo}$  & Ground truth & Zhang et al.~\cite{zhang2018single} & BDN~\cite{eccv18refrmv_BDN} \\

\includegraphics[width=0.1601\linewidth]{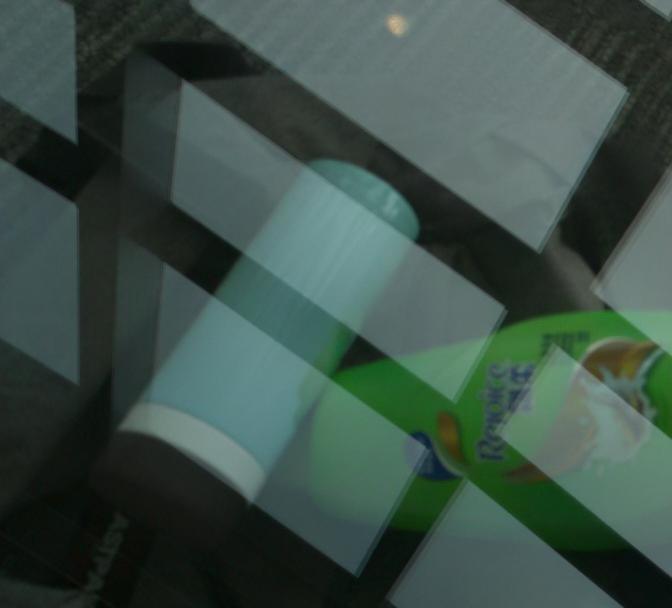}&
\includegraphics[width=0.1601\linewidth]{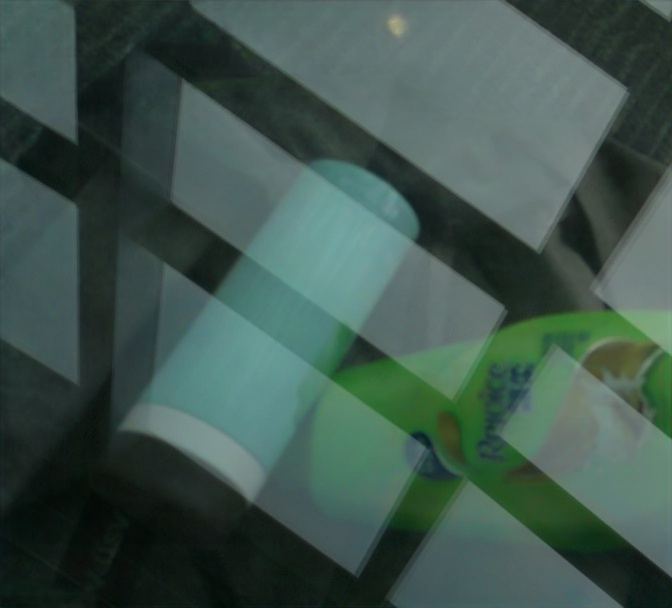}&
\includegraphics[width=0.1601\linewidth]{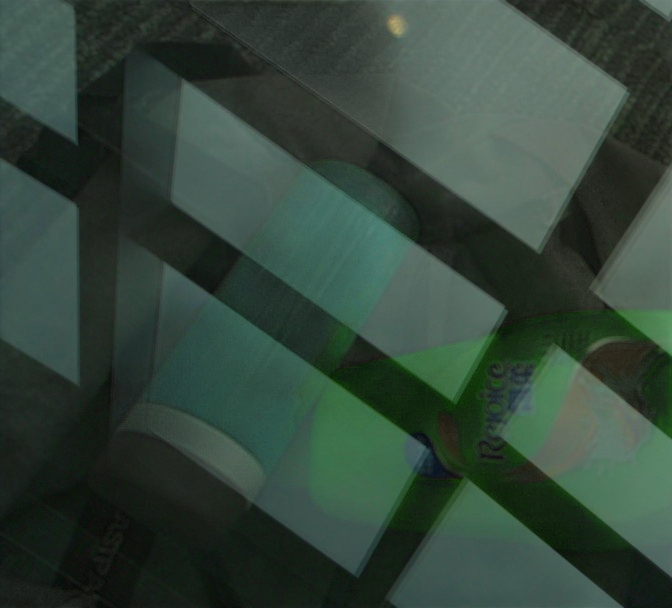}&
\includegraphics[width=0.1601\linewidth]{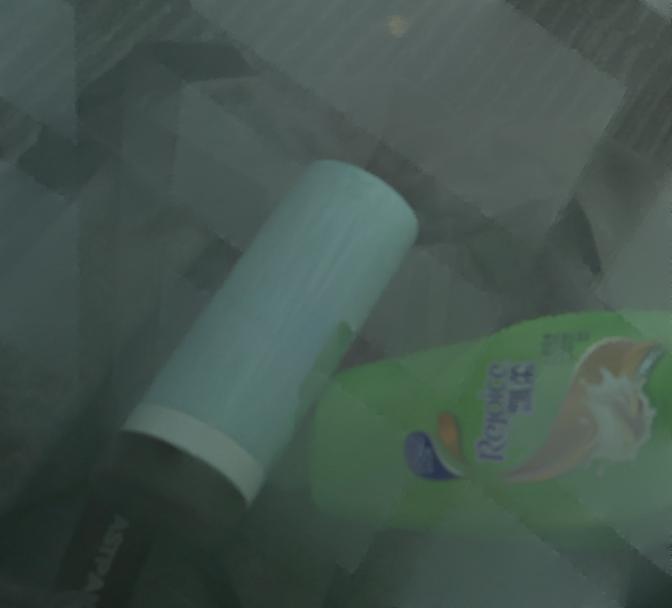}&
\includegraphics[width=0.1601\linewidth]{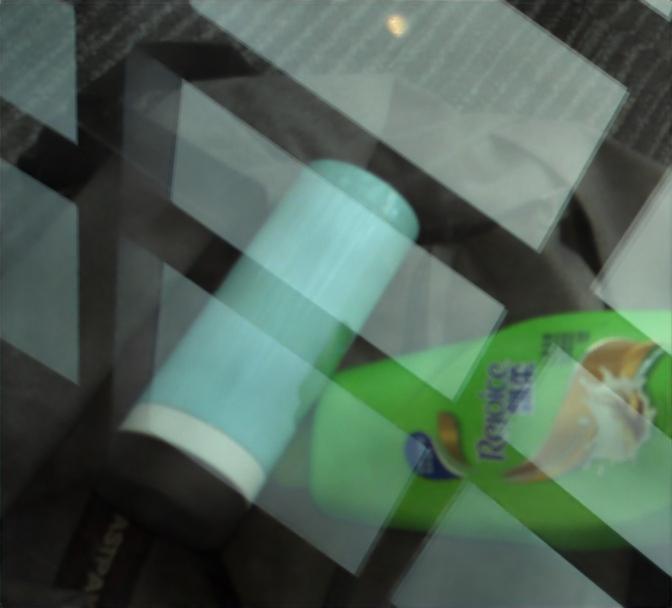}&
\includegraphics[width=0.1601\linewidth]{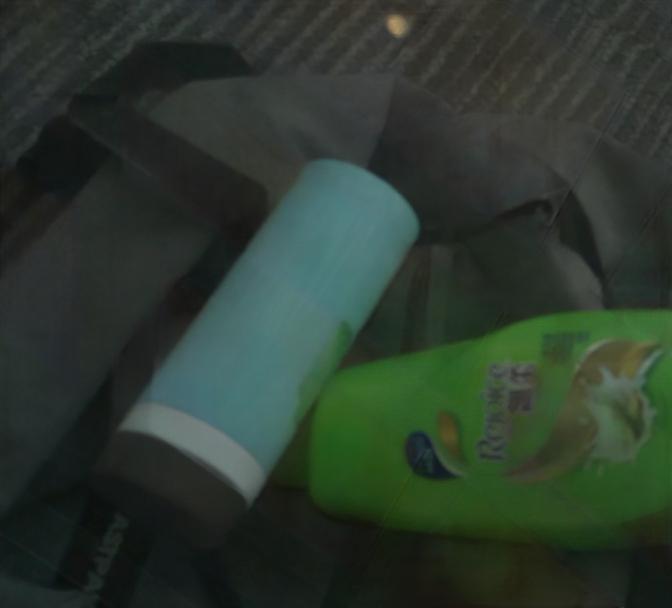}\\

Wei et al.~\cite{wei2019single_ERR} &  Kim et al.~\cite{Kim_2020_CVPR}  & Li et al.~\cite{Li_2020_CVPR} &{Agrawal et al.~\cite{agrawal2005removing_flash}} & SDN~\cite{chang2020siamese}&Ours\\
\end{tabular}
\vspace{1mm}
\caption{Qualitative comparison to baselines on a real-world image that contains strong reflection.}
\label{fig:StrongComparison}
\end{figure*}

\begin{figure*}[]
\centering
\begin{tabular}{@{}c@{\hspace{1mm}}c@{\hspace{1mm}}c@{\hspace{1mm}}c@{\hspace{1mm}}c@{}}

\includegraphics[width=0.193\linewidth]{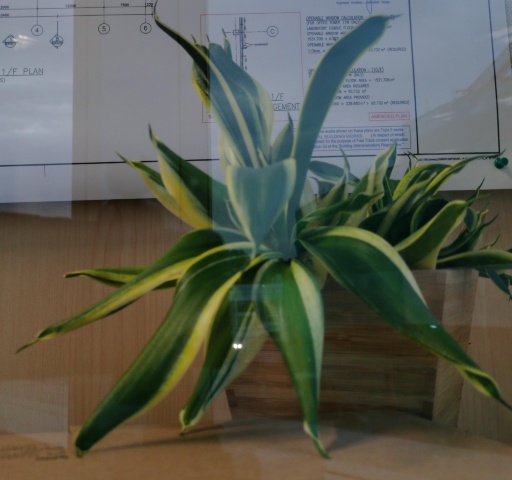}&
\includegraphics[width=0.193\linewidth]{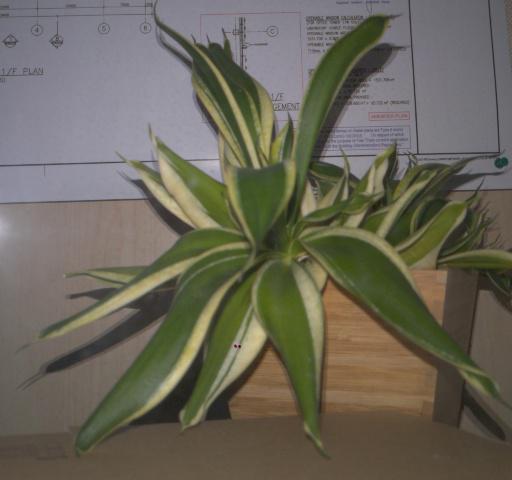}&
\includegraphics[width=0.193\linewidth]{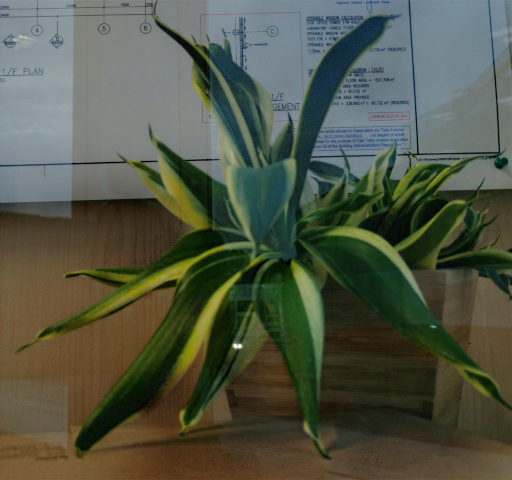}&
\includegraphics[width=0.193\linewidth]{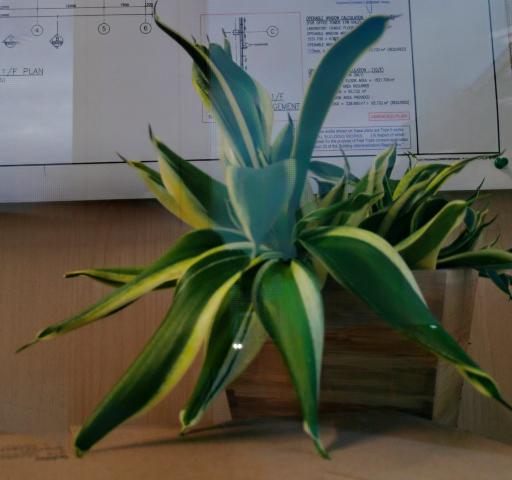}&
\includegraphics[width=0.193\linewidth]{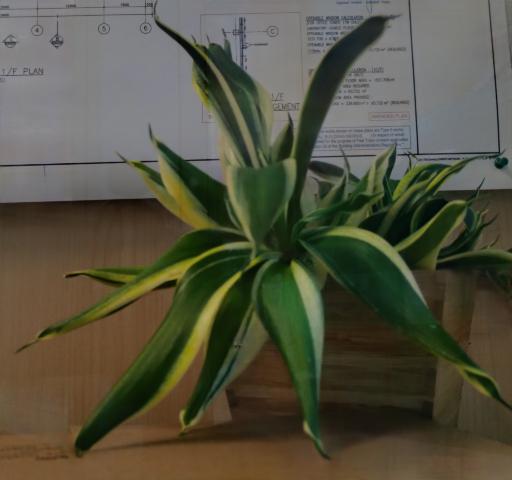}\\

\includegraphics[width=0.193\linewidth]{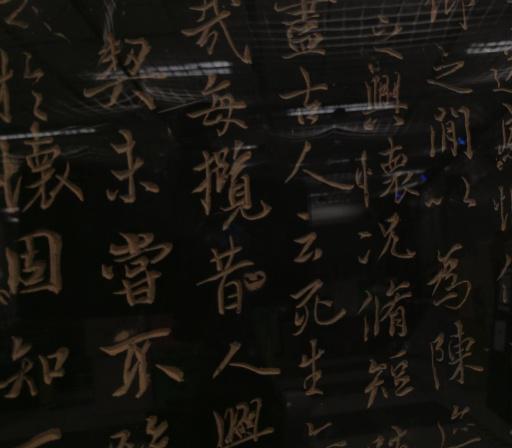}&
\includegraphics[width=0.193\linewidth]{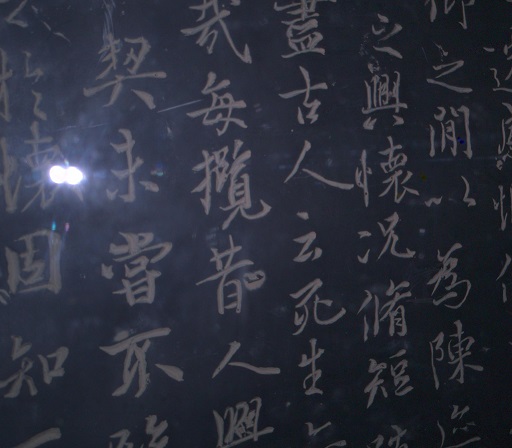}&
\includegraphics[width=0.193\linewidth]{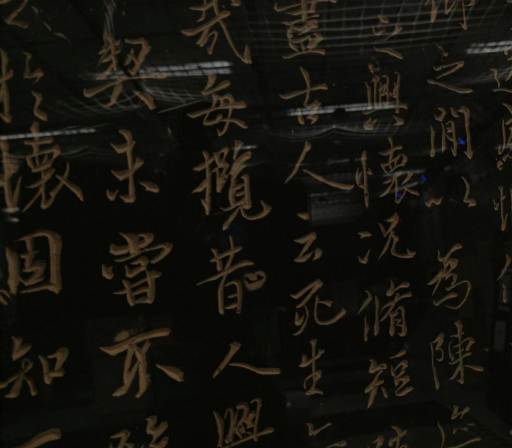}&
\includegraphics[width=0.193\linewidth]{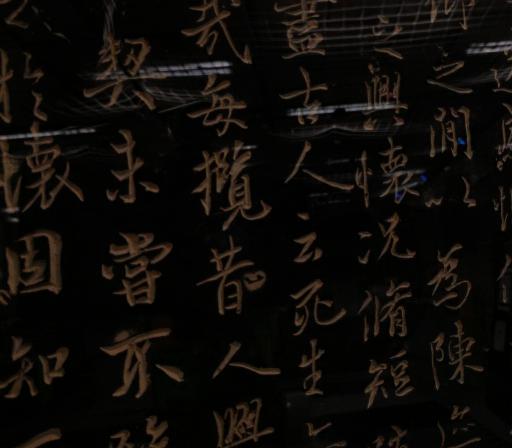}&
\includegraphics[width=0.193\linewidth]{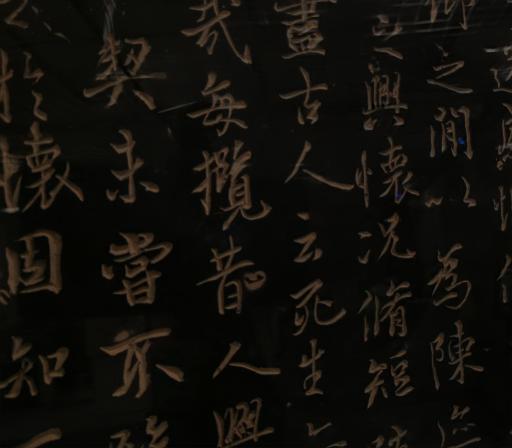}\\

Input ambient image  &Processed $I_{fo}$  & Li et al.~\cite{Li_2020_CVPR}& Wei et al.~\cite{wei2019single_ERR} & Ours\\
\end{tabular}
\vspace{1mm}
\caption{Qualitative comparison to single image based baselines~\cite{Li_2020_CVPR,wei2019single_ERR} on real-world images.}
\label{fig:ComparisonSingle}
\end{figure*}

\begin{figure}
\centering
\begin{tabular}{@{}c@{\hspace{1mm}}c@{\hspace{1mm}}c@{\hspace{1mm}}c@{}}
&
\includegraphics[width=0.321\linewidth]{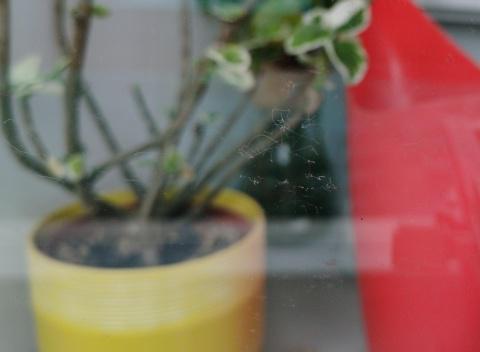}&
\includegraphics[width=0.321\linewidth]{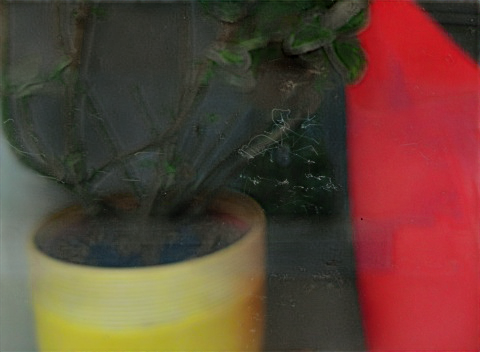}&
\includegraphics[width=0.321\linewidth]{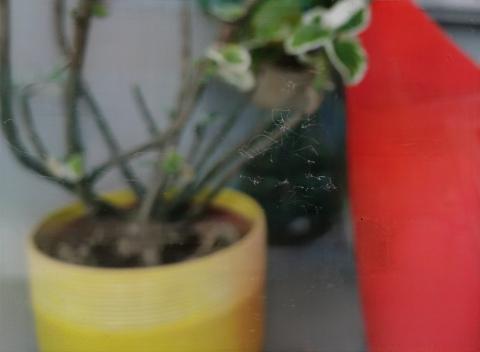}\\
&Input $I_{a}$ & Li et al.~\cite{Li_2020_CVPR} &  Ours
\end{tabular}
\vspace{1mm}
\caption{Comparison to Li et al.~\cite{Li_2020_CVPR} on a real-world image that contains blurry transmission. }
\label{fig:ComparisonBlur}
\end{figure}

\subsection{Limitation}
The reflection-free cue is based on the quality of flash-only image. 
If all objects in transmission are too far and are not illuminated by the flash, there would be no difference between the ambient image and the flash image (i.e., the flash-only image will be totally black except for reflected flash) due to the irradiance falloff problem in Eq.~\ref{eq:Falloff}. In this case, our model will be degraded to single image reflection removal. Also, if the objects in transmission move rapidly (larger motion within the exposure time), the flash-only image can have a serious misalignment problem. We believe other methods should be proposed to solve these cases well.

\section{Flash-only Reflection Removal Dataset}
\label{subsec:Dataset}

\textbf{Real-world data.}
Our method requires a pair of raw flash/ambient images. Since there is no existing dataset, we construct the first real-world dataset that contains raw data for flash-based reflection removal. This dataset is collected by Nikon Z6 and a smartphone camera Huawei Mate30. We control the camera setting (e.g., exposure) to make sure that Eq.~\ref{eq:pureflash} holds. The collection procedure is as follows: 

1) Fix the focal length, aperture, exposure time, and ISO. 

2) Take the ambient image $I_a$ ($I_a^{raw}$). 

3) Turn on the flash and take the flash image $I_f$ ($I_f^{raw}$). 

4) To get the ground truth $T_a$, we turn off the flash and take an extra reflection image $R_a$ ($R_a^{raw}$). Note that this step is unnecessary at test time.

To collect high-quality data with perfect alignment, we use a tripod to fix the camera, as shown in Fig.~\ref{fig:App}(a). In practice, steps (1)-(3) can be programmed to be implemented automatically with a single shutter-press on mobile phones, as shown in Fig.~\ref{fig:App}(b). By doing so, the extra cost is mainly longer exposure time compared with single image methods. In the next section, we demonstrate that this extra flash image can robustly and effectively improve performance.

We collect ground truth ambient transmission $T_a$ for training and evaluation. Specifically, we obtain $T_a^{raw} = I_a^{raw} - R_a^{raw}$ in raw (linear) space~\cite{Lei_2020_CVPR}. Thus, an extra reflection image under ambient illumination is captured. Then, ISP is implemented for each raw image similar to processing pipeline in Sec.~\ref{sec:data_prepare}. We adopt the metadata of $I_a^{raw}$ to process the raw data $T_a^{raw}$. At last, we crop the area where the transmission is valid following Lei et al.~\cite{Lei_2020_CVPR}. Briefly speaking, we capture a set $\{I_a^{raw}, I_f^{raw}, R_a^{raw}\}$ 
and process these three images to get the set $\{I_a, I_f, I_{fo}, T_a, R_a\}$. In total, we collect 157 sets of real-world images. 

\textbf{Synthetic data.} Since the real-world dataset cannot provide enough data for training, we construct an extra synthetic dataset. We use 1964 ambient transmission images $T_a$ and flash-only transmission images $I_{fo}$ from a flash dataset~\cite{aksoy2018ECCV_flashdataset}. Two kinds of reflections $R_a$ are provided for each $T_a$ to synthesize the ambient image $I_a$. The first type of reflection is real-world reflections collected by Wan et al.~\cite{wan2019corrn}. Then, since there are many blurry reflections and few sharp reflections in their dataset~\cite{wan2019corrn}, we use an arbitrary ambient image that is quite sharp as the second type of reflection. We reverse gamma correction to mimic the raw data and synthesize $I_a^{raw}$ by $I_a^{raw}=R_a^{raw}+T_a^{raw}$. 

\textbf{Dataset split.}
For the real-world dataset, we use 77, 30, 50 sets of images for training, validation, and evaluation. There is no overlapping reflection or transmission between the training and test sets. The synthetic data is only used as a supplement for training since the real-world reflection images in CoRRN~\cite{wan2019corrn} are in a chaotic order, and no dataset split is available. 


\section{Experiments}

\subsection{Comparison to Baselines} 
We first select two flash-based reflection removal methods: Agrawal et al.~\cite{agrawal2005removing_flash} and SDN~\cite{chang2020siamese}. Then we select several single image methods for comparison, including Zhang et al.~\cite{zhang2018single}, Wei et al.~\cite{wei2019single_ERR}, BDN~\cite{eccv18refrmv_BDN}, Li et al.~\cite{Li_2020_CVPR}, and Kim et al.~\cite{Kim_2020_CVPR}. For Agrawal et al.~\cite{agrawal2005removing_flash}, we observe that it is wrongly used in the comparison of SDN~\cite{chang2020siamese}: they use the flash image instead of the flash-only image as guidance; in our comparison, we adopt the flash-only image as the input to Agrawal et al.~\cite{agrawal2005removing_flash}. For SDN~\cite{chang2020siamese}, we use predicted ambient transmission for quantitative comparison. We retrain the models whose training codes are available on our constructed training set and choose the better results between pretrained models and retrained models.

In Table~\ref{table:Metrics}, we adopt PSNR, SSIM, and LPIPS~\cite{zhang2018lpips} as quantitative evaluation metrics, and our model obtains the best scores on all metrics. Specifically, our method outperforms state-of-the-art reflection removal approaches by more than 5.23dB in PSNR, 0.04 in SSIM, and 0.068 in LPIPS on the real-world dataset. 

In Fig.~\ref{fig:StrongComparison}, we compare our approach with all mentioned baselines. Both single image baselines~\cite{Kim_2020_CVPR,Li_2020_CVPR,wei2019single_ERR,eccv18refrmv_BDN, zhang2018single} and flash-based baselines~\cite{agrawal2005removing_flash,chang2020siamese} cannot correctly remove reflection. As can be seen, our approach can remove very strong reflection and recover underlying transmission. It is because processed flash-only image $I_{fo}$ is still reflection-free for strong reflection, and thus provides strong guidance. 
In Fig.~\ref{fig:ComparisonSingle}, we further compare our method with single image methods~\cite{Li_2020_CVPR,wei2019single_ERR} that obtains quantitative scores. 
In the first row, the edge of reflection is sharp. The second row is a picture of calligraphy writing, in which both reflection and transmission have rare semantic information. As can be seen, two single-image methods~\cite{Li_2020_CVPR,wei2019single_ERR} cannot remove the reflections. 
Our method removes reflections well since the reflection-cue is independent of the appearance (e.g., smoothness and semantic information) of reflection. 

In Fig.~\ref{fig:ComparisonBlur}, we compare with Li et al.~\cite{Li_2020_CVPR} on an image that contains blurry transmission. The result of  Li et al.~\cite{Li_2020_CVPR} remove transmission wrongly since their method cannot distinguish the reflection correctly. As a comparison, our approach can easily distinguish the reflection and avoid removing transmission wrongly because the reflection-free cue is independent to smoothness.

\begin{figure*}
\centering
\begin{tabular}{@{}c@{\hspace{1mm}}c@{\hspace{1mm}}c@{\hspace{1mm}}c@{\hspace{1mm}}c@{\hspace{1mm}}c@{}}
&
\includegraphics[width=0.191\linewidth]{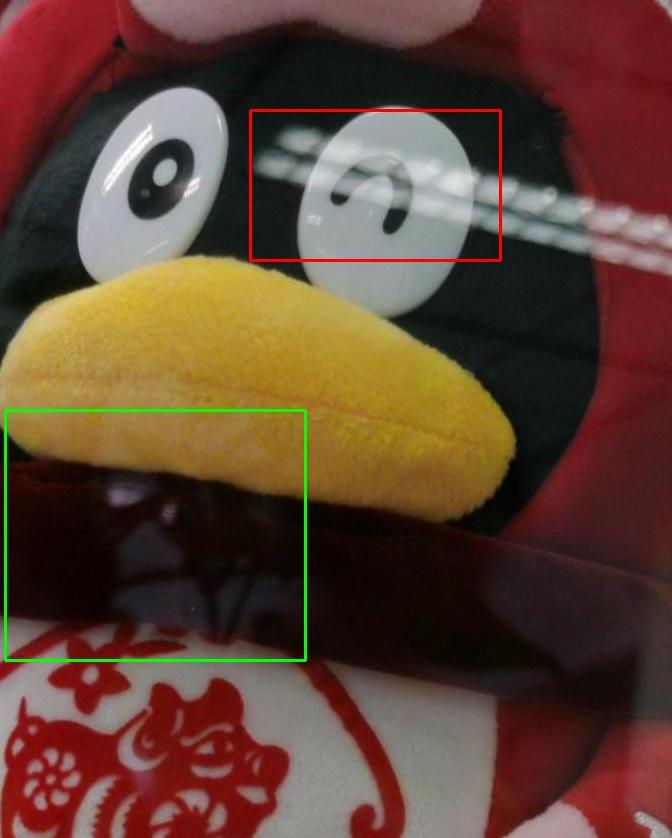}&
\includegraphics[width=0.191\linewidth]{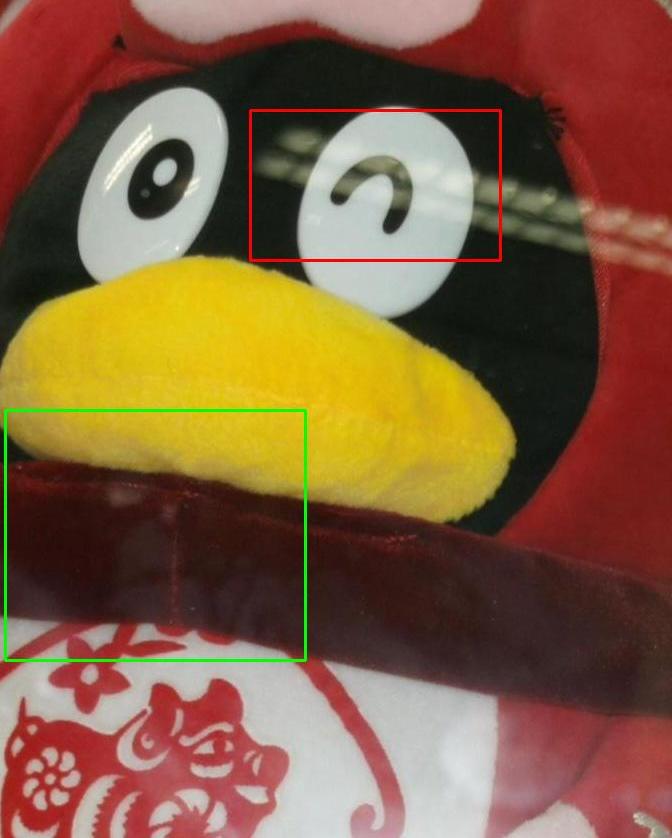}&
\includegraphics[width=0.191\linewidth]{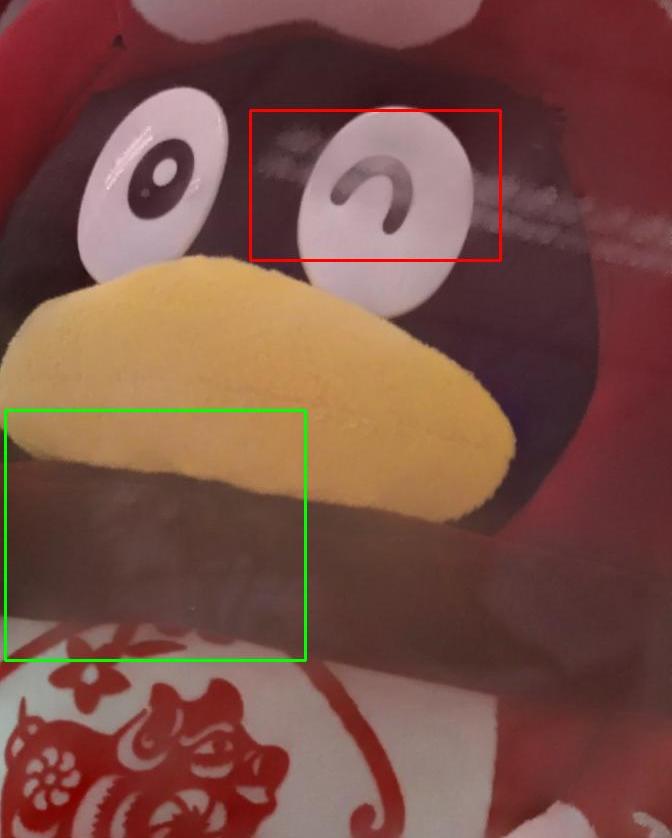}&
\includegraphics[width=0.191\linewidth]{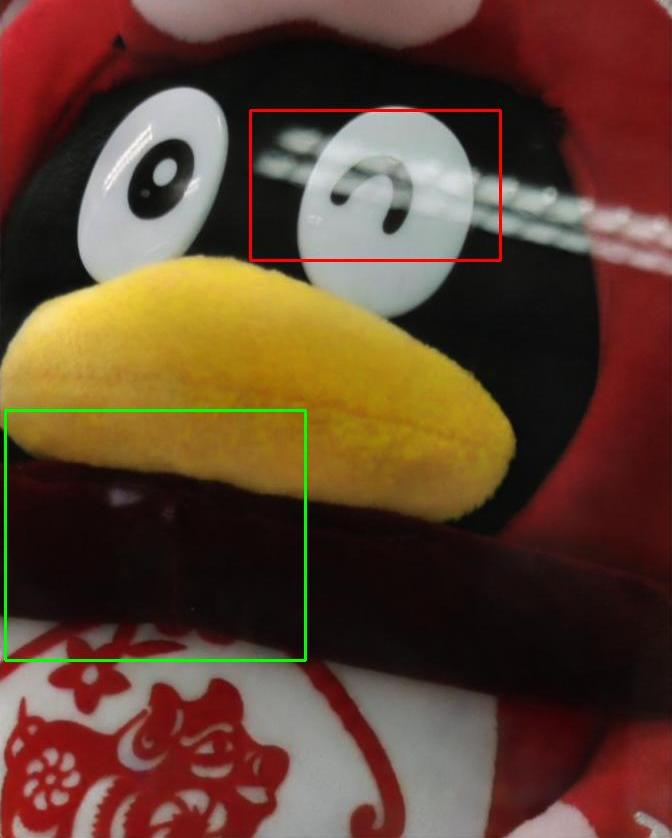}&
\includegraphics[width=0.191\linewidth]{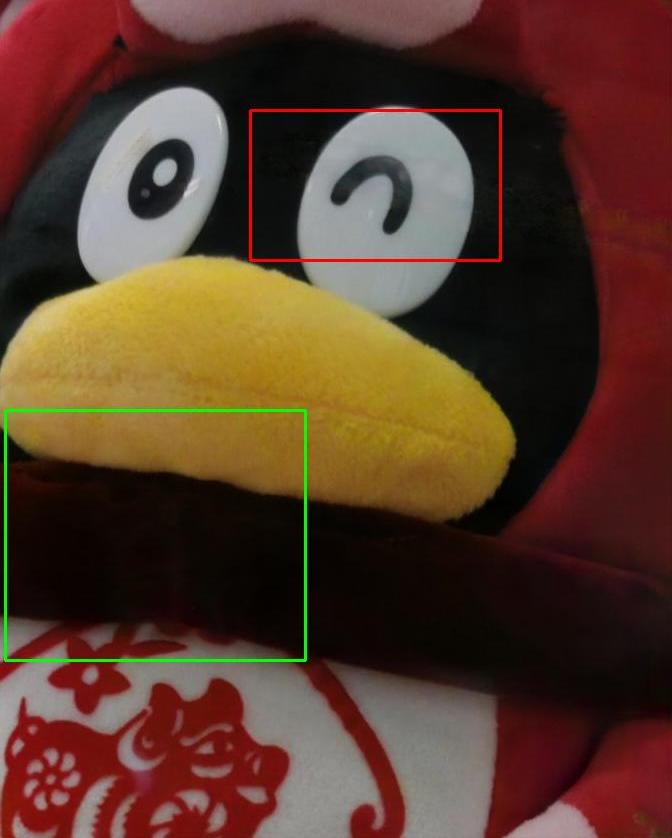}
\\
\rotatebox{90}{\small Weak reflection area}&
\includegraphics[width=0.191\linewidth]{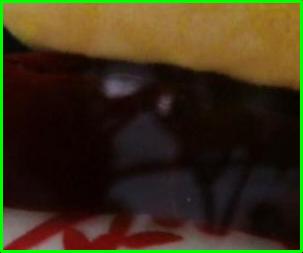}&
\includegraphics[width=0.191\linewidth]{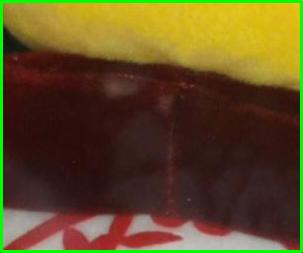}&
\includegraphics[width=0.191\linewidth]{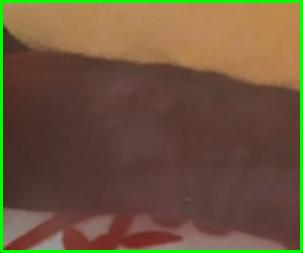}&
\includegraphics[width=0.191\linewidth]{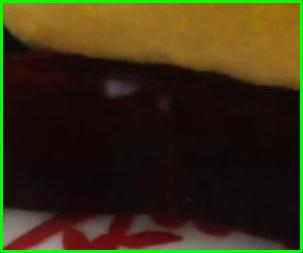}&
\includegraphics[width=0.191\linewidth]{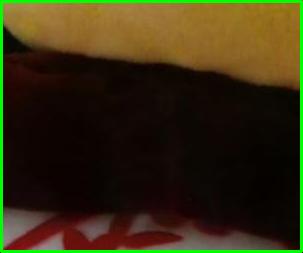}
\\

&Input ambient image & Input flash image & Agrawal et al.~\cite{agrawal2005removing_flash} & SDN~\cite{chang2020siamese}& Ours \\
\end{tabular}
\vspace{1mm}
\caption{Qualitative comparison to flash-based reflection removal baselines on real-world images. Results of Agrawal et al.~\cite{agrawal2005removing_flash} contain reflection residuals and are over-smooth. For SDN~\cite{chang2020siamese}, they can remove the weak reflection but cannot remove the strong reflection.}
\label{fig:ComparisonFlash}
\end{figure*}

In Fig.~\ref{fig:StrongComparison} and Fig.~\ref{fig:ComparisonFlash}, we compare our method with two flash-based methods~\cite{agrawal2005removing_flash, chang2020siamese}.
The results of Agrawal et al.~\cite{agrawal2005removing_flash} usually remove too many details and cannot completely remove reflection. For SDN~\cite{chang2020siamese}, they can remove weak reflection but cannot remove strong reflection. It is because they require the reflection is well suppressed, but the strong reflection cannot be suppressed by flash. Our method removes both weak and strong reflection. Also, the details and color are consistent with ambient images in our results.

\begin{figure*}
\centering
\begin{tabular}{@{}c@{\hspace{1mm}}c@{\hspace{1mm}}c@{\hspace{1mm}}c@{\hspace{1mm}}c@{\hspace{1mm}}c@{}}
&
\includegraphics[width=0.193\linewidth]{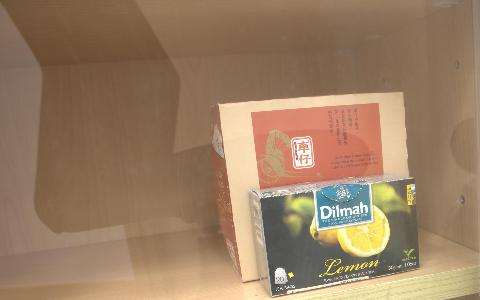}&
\includegraphics[width=0.193\linewidth]{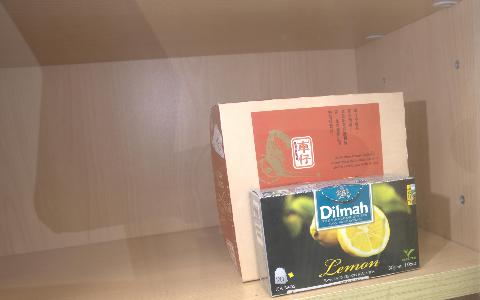}&
\includegraphics[width=0.193\linewidth]{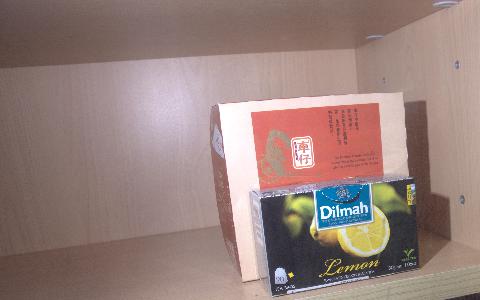}&
\includegraphics[width=0.193\linewidth]{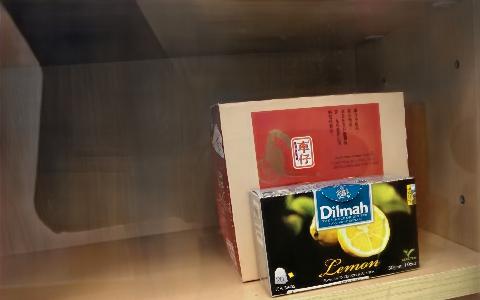}&
\includegraphics[width=0.193\linewidth]{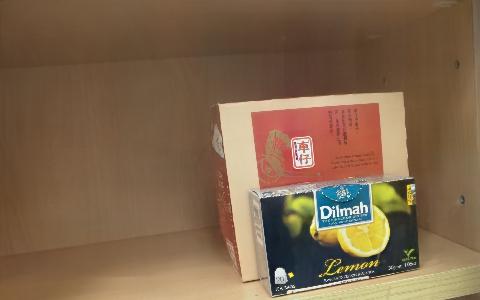}
\\

&
(a) Input $I_{a}$ &(b) Input $I_{f}$ & (c) Processed $I_{fo}$ & (d) Our $\hat T$: using $I_f$ & (e) Our $\hat T$: using $I_{fo}$  \\

\end{tabular}
\vspace{1mm}
\caption{Qualitative comparison between using $I_a+I_f$ and $I_a+I_{fo}$ as input. Note that the edge suppressed in flash image (b) is removed in (d). However, most reflections are not suppressed, and thus the perceptual quality of $(d)$ is not good.}
\label{fig:AblFlash}
\end{figure*}

\begin{table}[]
\small
\centering
\renewcommand{\arraystretch}{1.2}
\begin{tabular}{l@{\hspace{6mm}}lc@{\hspace{4mm}}c@{\hspace{4mm}}c@{\hspace{4mm}}}
\toprule[1pt]

Architecture& Input  &PSNR$\uparrow$ &  SSIM$\uparrow$ &  LPIPS$\downarrow$\\ 
\midrule
$g_B$& $I_a$+$I_{f}$  & 26.99 & 0.911 & 0.204       \\ 
$g_B$& $I_a$+$I_{fo}$ & \underline{27.55} & \underline{0.917} & \underline{0.187}  \\ 
$g_R + g_T$& $I_a$          & 25.13 & 0.888 & 0.258   \\ 
$g_R + g_T$& $I_a$+$I_{f}$  & 27.21 & \underline{0.917} & 0.196  \\ 
$g_R + g_T$& $I_a$+$I_{fo}$ & \textbf{29.76}  & \textbf{0.930} & \textbf{0.156}    \\ 
\bottomrule[1pt]
\end{tabular}
\vspace{1mm}
\caption{Quantitative comparison among our complete model and multiple ablated models of our methods on a real-world dataset. }
\label{table:Ablation study}
\end{table}

\subsection{Ablation Study} 

\textbf{Reflection-free cues.} 
Although it is quite simple to compute the reflection-free flash-only image, it can improves quantitative and qualitative results a lot. To demonstrate the importance of $I_{fo}$, we modify the input of the first network: (1) Replace $I_{fo}$ with $I_{f}$. (2) Use a single $I_a$ as input. Table~\ref{table:Ablation study} shows the quantitative results. Under the same training setting, using a single $I_a$ gets the worst scores, and replacing $I_{fo}$ with $I_f$ also degrades the performance. 

We find that the weakness of using the flash image $I_f$ instead of the flash-only image $I_{fo}$ is similar to SDN~\cite{chang2020siamese} in the qualitative comparison. In Fig.~\ref{fig:AblationStudy}, the reflection is not well suppressed by flash, but the flash-only image is still reflection-free. In this case, replacing $I_{fo}$ with $I_{f}$ performs poorly. Another example is also shown in Fig.~\ref{fig:AblFlash}(d), replacing $I_{fo}$ by $I_f$ leads to obvious artifacts when reflection cannot be suppressed by the flash. Moreover, it cannot handle novel shadows brought by the flash.

\textbf{Dedicated architecture.} 
As introduced in Sec.~\ref{subsec:basemodel}, the dedicated architecture is vital to avoid absorbing artifacts of flash-only images. In Fig.~\ref{fig:AblationStudy}, the base model can remove the reflection well, but artifacts (e.g., color distortion) appear in the result. As a comparison, the result of complete model does not contain obvious artifacts. In addition to, using our dedicated architecture also improves the quantitative performance a lot, as shown in Table~\ref{table:Ablation study}.

Note that although `$g_B$, $I_a + I_{fo}$' has similar quantitative scores with `$g_R+g_T, I_{a}+I_{f}$', the reasons for degradation are different: we observe the former can remove most reflection but usually has artifacts of flash-only images; the latter generally cannot remove strong reflection correctly.

\subsection{Discussion}
Flash-only images can be applied to various tasks. In addition to our reflection-free flash-only cues, we also notice other attractive cues exist in the flash-only images. For example, the shadow of the ambient image is invisible in the corresponding flash-only image. Namely, shadow-free flash-only cues also exist and they can be used for shadow removal~\cite{lu2006shadow}. Besides, there is only a single light source in the flash-only image, which is an important property to many tasks (such as photometric stereo~\cite{Chen_2019_CVPR}). We believe more applications of flash-only images are yet to be studied.

\section{Conclusion}
We propose a very simple yet effective cue called \emph{reflection-free cue} for reflection removal, which is independent of the appearance and strength of reflection. The reflection-free cue is based on the fact that objects in reflection do not directly receive light from the flash and the reflected flash is weak. With a reflection-free flash-only image as guidance, estimating the reflection becomes much easier. Since the flash-only image has obvious artifacts, we propose a dedicated architecture to avoid 
absorbing artifacts of flash-only images and utilize the cue better. As a result, our model outperforms state-of-the-art methods significantly on a real-world dataset. Also, the qualitative results show that our method can robustly remove various kinds of reflections. We also analyze the flash-based method's feasibility and find it simple to continuously take two images, making it practical in real-world applications. 

\section*{Acknowledgements}
\addcontentsline{toc}{section}{Acknowledgements} We thank Xuaner Zhang, Changlin Li, Yazhou Xing, and anonymous reviewers for helpful discussions on the paper. 


{\small
\bibliographystyle{ieee_fullname}
\bibliography{egbib}
}

\end{document}


\title{Robust Reflection Removal with Reflection-free Flash-only Cues\\\emph{Supplementary Material}}

\author{First Author\\
Institution1\\
Institution1 address\\
{\tt\small firstauthor@i1.org}
\and
Second Author\\
Institution2\\
First line of institution2 address\\
{\tt\small secondauthor@i2.org}
}

\maketitle

\begin{figure}
\centering
\begin{tabular}{@{}c@{}}
\includegraphics[width=1.0\linewidth]{Figure/supplement/flashonly_ill.pdf}\\
\end{tabular}
\caption{An illustration model of the formation of the transmission and reflection in the flash-only image. The light of transmission/reflection received by the camera is transmitted/reflected at least twice on the glass. 
The reflected flash is quite weak since the reflectance of glass is usually small.}
\label{fig:Illustration.}
\end{figure}

\section{Analysis for Reflection-free Cues}
\subsection{Reflectance of Glasses}
Fig.~\ref{fig:Illustration.} shows an illustration model. For objects in the reflection of ambient images, it can only receive the flash reflected from the glass under flash-only illumination. The objects can then reflect the light to the glass, and thus the camera can receive the light. Briefly speaking, the light of reflection received by the camera is reflected at least twice on the glass. Similarly, the light of transmission received by the camera is transmitted at least twice on the glass. 

Let $t\in [0,1]$ and $r\in[0,1]$ be the transmittance and reflectance of glass. For the same scene and same flash, since the light of transmission/reflection received by the camera is transmitted/reflected at least \emph{twice} on the glass, we have:
\begin{align}
    R_{fo} & \propto r^2, \\
    T_{fo} & \propto t^2,
\end{align}
where $R_{fo}$ and $T_{fo}$ is the reflection image and transmission image under flash-only illumination. 

Glass usually has quite small reflectance $r$, e.g. $10\%$. If we do not consider the absorbtion effect of glass, we have:
\begin{align}
    r+t=1.
\end{align}
We can simply get $r^2=0.01$ and $t^2=0.81$. Hence, the reflection usually is too weak to be visible.

\subsection{Distance between Objects and Glasses}
We need to consider the distance between the object and the flash light due to the irradiance falloff. Since the distance between flash and glass is usually small, we adopt the position of glass as the position of the flash. Let $d_t, d_r$ be the distance to the glass of transmission and reflection. Similarly, we have:
\begin{align}
    R_{fo} & \propto \frac{1}{d_r^2},  \\
    T_{fo} & \propto \frac{1}{d_t^2}.
\end{align}

We observe that the objects in the reflection of ambient images are usually far from the glass. Hence, the reflection in the flash-only image is further weakened.

\section{Reflection-free Flash-only Images}
In Fig.~\ref{fig:fo_samples}, we present examples of flash-only images and their corresponding ambient/flash images. The quality of flash-only images is decided by the objects in transmission and the flash light. 

In Fig.~\ref{fig:fo_samples} (a), we can artifacts in the flash-only image, including highlight, shadow. However, in the flash-only image, the reflection totally disappears. 

In Fig.~\ref{fig:fo_samples} (b), we can see the reflection on a poster. The reflection is strong, and it is not well suppressed in the flash image. However, in the flash-only image, the reflection totally disappears. 

For overexposure area in the ambient image or flash-image, $I_f^{raw} - I_a^{raw}$ can lead to abnormal values. As shown in some pink area in Fig.~\ref{fig:fo_samples} (b) and Fig.~\ref{fig:fo_samples} (c).

In Fig.~\ref{fig:fo_samples} (d) and Fig.~\ref{fig:fo_samples} (e), we can see artifacts due to slight misalignment (zoom in to check the details).  Although, in Fig.~\ref{fig:fo_samples} (e), the dust and scracthes on glass on illuminated.

Although these images have various artifacts, they have a common property that the flash-only image is almost totally reflection-free.

\begin{figure*}
\centering
\begin{tabular}{@{}c@{\hspace{1mm}}c@{\hspace{1mm}}c@{\hspace{1mm}}c@{}}
\rotatebox{90}{ \hspace{9mm}  (a) Sample 1}&
\includegraphics[width=0.31\linewidth]{Figure/supplement/flashonly/object_0010_ambient.jpg}&
\includegraphics[width=0.31\linewidth]{Figure/supplement/flashonly/object_0010_flash.jpg}&
\includegraphics[width=0.31\linewidth]{Figure/supplement/flashonly/object_0010_pureflash.jpg}\\

\rotatebox{90}{ \hspace{9mm} (b) Sample 2}&
\includegraphics[width=0.31\linewidth]{Figure/supplement/flashonly/object_0000_ambient.jpg}&
\includegraphics[width=0.31\linewidth]{Figure/supplement/flashonly/object_0000_flash.jpg}&
\includegraphics[width=0.31\linewidth]{Figure/supplement/flashonly/object_0000_pureflash.jpg}\\

\rotatebox{90}{ \hspace{9mm} (c) Sample 3 }&
\includegraphics[width=0.31\linewidth]{Figure/supplement/flashonly/object_0009_ambient.jpg}&
\includegraphics[width=0.31\linewidth]{Figure/supplement/flashonly/object_0009_flash.jpg}&
\includegraphics[width=0.31\linewidth]{Figure/supplement/flashonly/object_0009_pureflash.jpg}\\

\rotatebox{90}{ \hspace{9mm} (d) Sample 4 }&
\includegraphics[width=0.31\linewidth]{Figure/supplement/flashonly/object_0001_ambient.jpg}&
\includegraphics[width=0.31\linewidth]{Figure/supplement/flashonly/object_0001_flash.jpg}&
\includegraphics[width=0.31\linewidth]{Figure/supplement/flashonly/object_0001_pureflash.jpg}\\

\rotatebox{90}{ \hspace{9mm} (e) Sample 5 }&
\includegraphics[width=0.31\linewidth]{Figure/supplement/flashonly/object_0005_ambient.jpg}&
\includegraphics[width=0.31\linewidth]{Figure/supplement/flashonly/object_0005_flash.jpg}&
\includegraphics[width=0.31\linewidth]{Figure/supplement/flashonly/object_0005_pureflash.jpg}\\

 &Ambient images & Flash images  & Flash-only images \\

\end{tabular}
\caption{Examples of reflection-free flash-only images. Flash-only images are visually reflection-free but have many artifacts.}
\label{fig:fo_samples}
\end{figure*}

\section{Perceptual Results}
\begin{figure*}[t!]
\centering
\begin{tabular}{@{}c@{\hspace{1mm}}c@{\hspace{1mm}}c@{\hspace{1mm}}c@{\hspace{1mm}}c@{}}
\includegraphics[width=0.2799\linewidth]{Figure/supplement/perceptual/set_0064/Input.jpg}&
\includegraphics[width=0.2799\linewidth]{Figure/supplement/perceptual/set_0064/flash.jpg}&
\includegraphics[width=0.2799\linewidth]{Figure/supplement/perceptual/set_0064/object_0032_pureflash.jpg}\\

Input $I_a$  &  Input $I_f$  & Processed $I_{fo}$   \\

\includegraphics[width=0.2799\linewidth]{Figure/supplement/perceptual/set_0064/Zhang.png}&
\includegraphics[width=0.2799\linewidth]{Figure/supplement/perceptual/set_0064/BDN.jpg}&
\includegraphics[width=0.2799\linewidth]{Figure/supplement/perceptual/set_0064/Wei.jpg}\\
Zhang et al.~\cite{zhang2018single}  &  BDN~\cite{eccv18refrmv_BDN}  &  Wei et al.~\cite{wei2019single_ERR}\\

\includegraphics[width=0.2799\linewidth]{Figure/supplement/perceptual/set_0064/Kim.jpg}&
\includegraphics[width=0.2799\linewidth]{Figure/supplement/perceptual/set_0064/Li.png}&
\includegraphics[width=0.2799\linewidth]{Figure/supplement/perceptual/set_0064/Chang.jpg}\\
Kim et al.~\cite{Kim_2020_CVPR}  &  Li et al.~\cite{Li_2020_CVPR}  & SDN~\cite{chang2020siamese}   \\

\includegraphics[width=0.2799\linewidth]{Figure/supplement/perceptual/set_0064/Ag.jpg}&
\includegraphics[width=0.2799\linewidth]{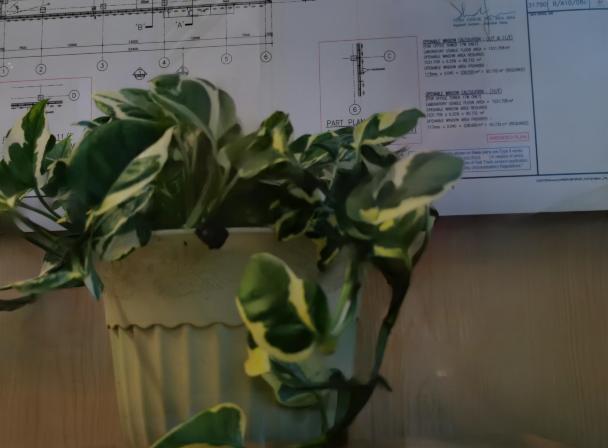}&
\\
Agrawal et al.~\cite{agrawal2005removing_flash}  &  Ours  &   \\

\end{tabular}
\vspace{1mm}
\caption{Qualitative comparison to baselines on a real-world image.}
\label{fig:perceptual_2}
\end{figure*}

\begin{figure*}[t!]
\centering
\begin{tabular}{@{}c@{\hspace{1mm}}c@{\hspace{1mm}}c@{\hspace{1mm}}c@{\hspace{1mm}}c@{}}
\includegraphics[width=0.2799\linewidth]{Figure/supplement/perceptual/set_0149/Input.jpg}&
\includegraphics[width=0.2799\linewidth]{Figure/supplement/perceptual/set_0149/flash.jpg}&
\includegraphics[width=0.2799\linewidth]{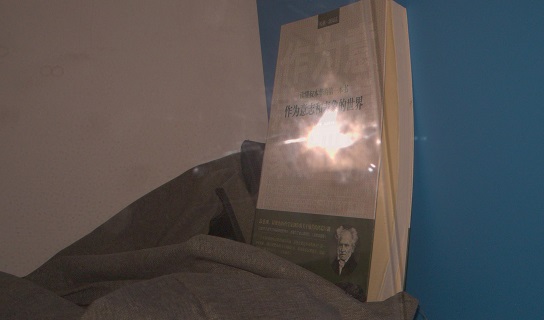}\\
Input $I_a$  &  Input $I_f$  & Processed $I_{fo}$   \\

\includegraphics[width=0.2799\linewidth]{Figure/supplement/perceptual/set_0149/Zhang.png}&
\includegraphics[width=0.2799\linewidth]{Figure/supplement/perceptual/set_0149/BDN.jpg}&
\includegraphics[width=0.2799\linewidth]{Figure/supplement/perceptual/set_0149/Wei.jpg}\\
Zhang et al.~\cite{zhang2018single}  &  BDN~\cite{eccv18refrmv_BDN}  &  Wei et al.~\cite{wei2019single_ERR}\\

\includegraphics[width=0.2799\linewidth]{Figure/supplement/perceptual/set_0149/Kim.jpg}&
\includegraphics[width=0.2799\linewidth]{Figure/supplement/perceptual/set_0149/Li.png}&
\includegraphics[width=0.2799\linewidth]{Figure/supplement/perceptual/set_0149/Chang.jpg}\\
Kim et al.~\cite{Kim_2020_CVPR}  &  Li et al.~\cite{Li_2020_CVPR}  & SDN~\cite{chang2020siamese}   \\

\includegraphics[width=0.2799\linewidth]{Figure/supplement/perceptual/set_0149/Ag.jpg}&
\includegraphics[width=0.2799\linewidth]{Figure/supplement/perceptual/set_0149/Ours.jpg}&
\\
Agrawal et al.~\cite{agrawal2005removing_flash}  &  Ours  &  \\

\end{tabular}
\vspace{1mm}
\caption{Qualitative comparison to baselines on a real-world image.}
\label{fig:perceptual_4}
\end{figure*}

We present more perceptual comparison to all baselines in Fig.~\ref{fig:perceptual_1}, Fig.~\ref{fig:perceptual_2}, Fig.~\ref{fig:perceptual_3}, and Fig.~\ref{fig:perceptual_4} on our real-world dataset. We also present perceptual comparison on our constructed synthetic, as shown in Fig.~\ref{fig:fake_1}, Fig.~\ref{fig:fake_2}, Fig.~\ref{fig:fake_3}, and Fig.~\ref{fig:fake_4}. Single image methods usually cannot remove the reflection correctly~\cite{Li_2020_CVPR,Kim_2020_CVPR,wei2019single_ERR,zhang2018single} or remove the transmission wrongly~\cite{eccv18refrmv_BDN}.
The results of Agrawal et al.~\cite{agrawal2005removing_flash} usually remove too many details and cannot completely remove the reflection. For SDN~\cite{chang2020siamese}, they can remove weak reflection but cannot remove the strong reflection. Our approach can accurately and robustly remove various reflections. Also, the details and color are consistent with ambient images in our results.

\begin{figure*}[t!]
\centering
\begin{tabular}{@{}c@{\hspace{1mm}}c@{\hspace{1mm}}c@{\hspace{1mm}}c@{\hspace{1mm}}c@{}}
\includegraphics[width=0.2799\linewidth]{Figure/supplement/perceptual/set_0053/Input.jpg}&
\includegraphics[width=0.2799\linewidth]{Figure/supplement/perceptual/set_0053/flash.jpg}&
\includegraphics[width=0.2799\linewidth]{Figure/supplement/perceptual/set_0053/flashonly.jpg}\\
Input $I_a$  &  Input $I_f$  & Processed $I_{fo}$   \\

\includegraphics[width=0.2799\linewidth]{Figure/supplement/perceptual/set_0053/Zhang.png}&
\includegraphics[width=0.2799\linewidth]{Figure/supplement/perceptual/set_0053/BDN.jpg}&
\includegraphics[width=0.2799\linewidth]{Figure/supplement/perceptual/set_0053/Wei.jpg}\\
Zhang et al.~\cite{zhang2018single}  &  BDN~\cite{eccv18refrmv_BDN}  &  Wei et al.~\cite{wei2019single_ERR}\\

\includegraphics[width=0.2799\linewidth]{Figure/supplement/perceptual/set_0053/Kim.jpg}&
\includegraphics[width=0.2799\linewidth]{Figure/supplement/perceptual/set_0053/Li.png}&
\includegraphics[width=0.2799\linewidth]{Figure/supplement/perceptual/set_0053/Chang.jpg}\\
Kim et al.~\cite{Kim_2020_CVPR}  &  Li et al.~\cite{Li_2020_CVPR}  & SDN~\cite{chang2020siamese}   \\

\includegraphics[width=0.2799\linewidth]{Figure/supplement/perceptual/set_0053/Ag.jpg}&
\includegraphics[width=0.2799\linewidth]{Figure/supplement/perceptual/set_0053/Ours.jpg}&
\\
Agrawal et al.~\cite{agrawal2005removing_flash}  &  Ours  &  \\

\end{tabular}
\vspace{1mm}
\caption{Qualitative comparison to baselines on a real-world image.}
\label{fig:perceptual_1}
\end{figure*}

\begin{figure*}[t!]
\centering
\begin{tabular}{@{}c@{\hspace{1mm}}c@{\hspace{1mm}}c@{\hspace{1mm}}c@{\hspace{1mm}}c@{}}
\includegraphics[width=0.2799\linewidth]{Figure/supplement/perceptual/set_0102/Input.jpg}&
\includegraphics[width=0.2799\linewidth]{Figure/supplement/perceptual/set_0102/flash.jpg}&
\includegraphics[width=0.2799\linewidth]{Figure/supplement/perceptual/set_0102/flashonly.jpg}\\
Input $I_a$  &  Input $I_f$  & Processed $I_{fo}$   \\

\includegraphics[width=0.2799\linewidth]{Figure/supplement/perceptual/set_0102/Zhang.png}&
\includegraphics[width=0.2799\linewidth]{Figure/supplement/perceptual/set_0102/BDN.jpg}&
\includegraphics[width=0.2799\linewidth]{Figure/supplement/perceptual/set_0102/Wei.jpg}\\
Zhang et al.~\cite{zhang2018single}  &  BDN~\cite{eccv18refrmv_BDN}  &  Wei et al.~\cite{wei2019single_ERR}\\

\includegraphics[width=0.2799\linewidth]{Figure/supplement/perceptual/set_0102/Kim.jpg}&
\includegraphics[width=0.2799\linewidth]{Figure/supplement/perceptual/set_0102/Li.png}&
\includegraphics[width=0.2799\linewidth]{Figure/supplement/perceptual/set_0102/Chang.jpg}\\
Kim et al.~\cite{Kim_2020_CVPR}  &  Li et al.~\cite{Li_2020_CVPR}  & SDN~\cite{chang2020siamese}   \\

\includegraphics[width=0.2799\linewidth]{Figure/supplement/perceptual/set_0102/Ag.jpg}&
\includegraphics[width=0.2799\linewidth]{Figure/supplement/perceptual/set_0102/ours_lp.jpg}&
\\
Agrawal et al.~\cite{agrawal2005removing_flash}  &  Ours  &  \\

\end{tabular}
\vspace{1mm}
\caption{Qualitative comparison to baselines on a real-world image.}
\label{fig:perceptual_3}
\end{figure*}

\begin{figure*}[t!]
\centering
\begin{tabular}{@{}c@{\hspace{1mm}}c@{\hspace{1mm}}c@{\hspace{1mm}}c@{\hspace{1mm}}c@{}}
\includegraphics[width=0.2799\linewidth]{Figure/supplement/perceptual/fake_0308/Input.jpg}&
\includegraphics[width=0.2799\linewidth]{Figure/supplement/perceptual/fake_0308/flash.jpg}&
\includegraphics[width=0.2799\linewidth]{Figure/supplement/perceptual/fake_0308/flashonly.jpg}\\
Input $I_a$  &  Input $I_f$  & Processed $I_{fo}$   \\

\includegraphics[width=0.2799\linewidth]{Figure/supplement/perceptual/fake_0308/Zhang.png}&
\includegraphics[width=0.2799\linewidth]{Figure/supplement/perceptual/fake_0308/BDN.jpg}&
\includegraphics[width=0.2799\linewidth]{Figure/supplement/perceptual/fake_0308/Wei.jpg}\\
Zhang et al.~\cite{zhang2018single}  &  BDN~\cite{eccv18refrmv_BDN}  &  Wei et al.~\cite{wei2019single_ERR}\\

\includegraphics[width=0.2799\linewidth]{Figure/supplement/perceptual/fake_0308/Kim.jpg}&
\includegraphics[width=0.2799\linewidth]{Figure/supplement/perceptual/fake_0308/Li.png}&
\includegraphics[width=0.2799\linewidth]{Figure/supplement/perceptual/fake_0308/Chang.jpg}\\
Kim et al.~\cite{Kim_2020_CVPR}  &  Li et al.~\cite{Li_2020_CVPR}  & SDN~\cite{chang2020siamese}   \\

\includegraphics[width=0.2799\linewidth]{Figure/supplement/perceptual/fake_0308/Ag.jpg}&
\includegraphics[width=0.2799\linewidth]{Figure/supplement/perceptual/fake_0308/Ours.png}&
\\
Agrawal et al.~\cite{agrawal2005removing_flash}  &  Ours  &  \\

\end{tabular}
\vspace{1mm}
\caption{Qualitative comparison to baselines on a synthetic image from Aksoy et al.~\cite{aksoy2018ECCV_flashdataset}}
\label{fig:fake_1}
\end{figure*}

\begin{figure*}[t!]
\centering
\begin{tabular}{@{}c@{\hspace{1mm}}c@{\hspace{1mm}}c@{\hspace{1mm}}c@{\hspace{1mm}}c@{}}
\includegraphics[width=0.2799\linewidth]{Figure/supplement/perceptual/fake_0048/Input.jpg}&
\includegraphics[width=0.2799\linewidth]{Figure/supplement/perceptual/fake_0048/flash.jpg}&
\includegraphics[width=0.2799\linewidth]{Figure/supplement/perceptual/fake_0048/flashonly.jpg}\\
Input $I_a$  &  Input $I_f$  & Processed $I_{fo}$   \\

\includegraphics[width=0.2799\linewidth]{Figure/supplement/perceptual/fake_0048/Zhang.png}&
\includegraphics[width=0.2799\linewidth]{Figure/supplement/perceptual/fake_0048/BDN.jpg}&
\includegraphics[width=0.2799\linewidth]{Figure/supplement/perceptual/fake_0048/Wei.jpg}\\
Zhang et al.~\cite{zhang2018single}  &  BDN~\cite{eccv18refrmv_BDN}  &  Wei et al.~\cite{wei2019single_ERR}\\

\includegraphics[width=0.2799\linewidth]{Figure/supplement/perceptual/fake_0048/Kim.jpg}&
\includegraphics[width=0.2799\linewidth]{Figure/supplement/perceptual/fake_0048/Li.png}&
\includegraphics[width=0.2799\linewidth]{Figure/supplement/perceptual/fake_0048/Chang.jpg}\\
Kim et al.~\cite{Kim_2020_CVPR}  &  Li et al.~\cite{Li_2020_CVPR}  & SDN~\cite{chang2020siamese}   \\

\includegraphics[width=0.2799\linewidth]{Figure/supplement/perceptual/fake_0048/Ag.jpg}&
\includegraphics[width=0.2799\linewidth]{Figure/supplement/perceptual/fake_0048/Ours.png}&
\\
Agrawal et al.~\cite{agrawal2005removing_flash}  &  Ours  &  \\

\end{tabular}
\vspace{1mm}
\caption{Qualitative comparison to baselines on a synthetic image from Aksoy et al.~\cite{aksoy2018ECCV_flashdataset}}
\label{fig:fake_2}
\end{figure*}

\begin{figure*}[t!]
\centering
\begin{tabular}{@{}c@{\hspace{1mm}}c@{\hspace{1mm}}c@{\hspace{1mm}}c@{\hspace{1mm}}c@{}}
\includegraphics[width=0.2799\linewidth]{Figure/supplement/perceptual/fake_0244/Input.jpg}&
\includegraphics[width=0.2799\linewidth]{Figure/supplement/perceptual/fake_0244/flash.jpg}&
\includegraphics[width=0.2799\linewidth]{Figure/supplement/perceptual/fake_0244/flashonly.jpg}\\
Input $I_a$  &  Input $I_f$  & Processed $I_{fo}$   \\

\includegraphics[width=0.2799\linewidth]{Figure/supplement/perceptual/fake_0244/Zhang.png}&
\includegraphics[width=0.2799\linewidth]{Figure/supplement/perceptual/fake_0244/BDN.jpg}&
\includegraphics[width=0.2799\linewidth]{Figure/supplement/perceptual/fake_0244/Wei.jpg}\\
Zhang et al.~\cite{zhang2018single}  &  BDN~\cite{eccv18refrmv_BDN}  &  Wei et al.~\cite{wei2019single_ERR}\\

\includegraphics[width=0.2799\linewidth]{Figure/supplement/perceptual/fake_0244/Kim.jpg}&
\includegraphics[width=0.2799\linewidth]{Figure/supplement/perceptual/fake_0244/Li.png}&
\includegraphics[width=0.2799\linewidth]{Figure/supplement/perceptual/fake_0244/Chang.jpg}\\
Kim et al.~\cite{Kim_2020_CVPR}  &  Li et al.~\cite{Li_2020_CVPR}  & SDN~\cite{chang2020siamese}   \\

\includegraphics[width=0.2799\linewidth]{Figure/supplement/perceptual/fake_0244/Ag.jpg}&
\includegraphics[width=0.2799\linewidth]{Figure/supplement/perceptual/fake_0244/Ours.png}&
\\
Agrawal et al.~\cite{agrawal2005removing_flash}  &  Ours  &  \\

\end{tabular}
\vspace{1mm}
\caption{Qualitative comparison to baselines on a synthetic image from Aksoy et al.~\cite{aksoy2018ECCV_flashdataset}}
\label{fig:fake_3}
\end{figure*}

\begin{figure*}[t!]
\centering
\begin{tabular}{@{}c@{\hspace{1mm}}c@{\hspace{1mm}}c@{\hspace{1mm}}c@{\hspace{1mm}}c@{}}
\includegraphics[width=0.2799\linewidth]{Figure/supplement/perceptual/fake_0076/Input.jpg}&
\includegraphics[width=0.2799\linewidth]{Figure/supplement/perceptual/fake_0076/flash.jpg}&
\includegraphics[width=0.2799\linewidth]{Figure/supplement/perceptual/fake_0076/flashonly.jpg}\\
Input $I_a$  &  Input $I_f$  & Processed $I_{fo}$   \\

\includegraphics[width=0.2799\linewidth]{Figure/supplement/perceptual/fake_0076/Zhang.png}&
\includegraphics[width=0.2799\linewidth]{Figure/supplement/perceptual/fake_0076/BDN.jpg}&
\includegraphics[width=0.2799\linewidth]{Figure/supplement/perceptual/fake_0076/Wei.jpg}\\
Zhang et al.~\cite{zhang2018single}  &  BDN~\cite{eccv18refrmv_BDN}  &  Wei et al.~\cite{wei2019single_ERR}\\

\includegraphics[width=0.2799\linewidth]{Figure/supplement/perceptual/fake_0076/Kim.jpg}&
\includegraphics[width=0.2799\linewidth]{Figure/supplement/perceptual/fake_0076/Li.png}&
\includegraphics[width=0.2799\linewidth]{Figure/supplement/perceptual/fake_0076/Chang.jpg}\\
Kim et al.~\cite{Kim_2020_CVPR}  &  Li et al.~\cite{Li_2020_CVPR}  & SDN~\cite{chang2020siamese}   \\

\includegraphics[width=0.2799\linewidth]{Figure/supplement/perceptual/fake_0076/Ag.jpg}&
\includegraphics[width=0.2799\linewidth]{Figure/supplement/perceptual/fake_0076/Ours.png}&
\\
Agrawal et al.~\cite{agrawal2005removing_flash}  &  Ours  &  \\

\end{tabular}
\vspace{1mm}
\caption{Qualitative comparison to baselines on a synthetic image from Aksoy et al.~\cite{aksoy2018ECCV_flashdataset}}
\label{fig:fake_4}
\end{figure*}

{\small
\bibliographystyle{ieee_fullname}
\bibliography{egbib}
}